%% file: BBOB2012_saACMES_noiseless.tex
\documentclass{sig-alternate}
\usepackage{graphicx}
\usepackage{tabularx}
\usepackage{xcolor}
\usepackage{float}
\usepackage{rotating}
\usepackage{xstring} 

\newcommand{\bbobdatapath}{ppdata/}
\input{\bbobdatapath bbob_pproc_commands.tex}
\graphicspath{{\bbobdatapath}}
 \newcommand{\algaperfprof}{BIPOP-CMA}
 \newcommand{\algbperfprof}{BIPOP-$^{s\ast}$aACM}
 \newcommand{\algcperfprof}{IPOP-aCMA}
 \newcommand{\algdperfprof}{IPOP-$^{\ast}$ACM}
% ...

\newcommand{\ERT}{\ensuremath{\mathrm{ERT}}}

\newcommand{\Df}{\ensuremath{\Delta f}}

\newcommand{\fopt}{\ensuremath{f_\mathrm{opt}}}
\newcommand{\ftarget}{\ensuremath{f_\mathrm{t}}}

\begin{document}

%
\conferenceinfo{GECCO'12,} {July 7--11, 2012, Philadelphia, Pennsylvania, USA.} 
\CopyrightYear{2012} 
\crdata{978-1-4503-1177-9/12/07} 
\clubpenalty=10000 
\widowpenalty = 10000 

\title{Black-box optimization benchmarking of IPOP-saACM-ES and BIPOP-saACM-ES on the BBOB-2012 noiseless testbed}

%
%
%
%
%

 \numberofauthors{3}
 \author{
 \alignauthor
 Ilya Loshchilov \\
 \affaddr{TAO, INRIA Saclay}\\
 \affaddr{U. Paris Sud, F-91405 Orsay}\\
 \and 
 \alignauthor
  Marc Schoenauer\\
 \affaddr{TAO, INRIA Saclay}\\
 \affaddr{U. Paris Sud, F-91405 Orsay}\\
 \email{firstname.lastname@inria.fr}
 \and 
 \alignauthor
  Mich\`ele Sebag\\
 \affaddr{CNRS, LRI UMR 8623}\\
 \affaddr{U. Paris Sud, F-91405 Orsay}\\
}

\def\SAM{$^{s\ast}$}
\def\IPOPsaACM{IPOP-\SAM\hspace{-0.50ex}aACM-ES}
\def\saACM{\SAM\hspace{-0.50ex}aACM-ES}
\def\sACM{\SAM\hspace{-0.50ex}ACM-ES}
\def\BIPOPsaACM{BIPOP-\SAM\hspace{-0.50ex}aACM-ES}
\def\IPOPCMA{IPOP-aCMA-ES}
\def\BIPOPCMA{BIPOP-CMA-ES}
\def\mulCMA{$(\mu / \mu_w , \lambda)$-CMA-ES}
\newcommand{\R}{\mathbb{R}}
\newcommand{\Rd}{\R^{D}}
\newcommand{\NormalNull}[1]{{\mathcal N}\hspace{-0.13em}\left(\ve{0},#1\right)}

\maketitle
\begin{abstract}
	In this paper, we study the performance of \IPOPsaACM\ and \BIPOPsaACM, recently proposed self-adaptive surrogate-assisted Covariance Matrix Adaptation Evolution Strategies.
	Both algorithms were tested using restarts till a total number of function evaluations of $10^6D$ was reached, where $D$ is the dimension of the function search space. 
	We compared surrogate-assisted algorithms with their surrogate-less versions \IPOPCMA\ and \BIPOPCMA, two algorithms with one of the best overall performance observed during the BBOB-2009 and BBOB-2010.
	
	The comparison shows that the surrogate-assisted versions outperform the original CMA-ES algorithms by a factor from 2 to 4 on 8 out of 24 noiseless benchmark problems, showing the best results among all algorithms of the BBOB-2009 and BBOB-2010 on Ellipsoid, Discus, Bent Cigar, Sharp Ridge and Sum of different powers functions.

\end{abstract}




\section{Introduction}

When dealing with expensive optimization objectives, the surrogate-assisted approaches proceed by learning
a surrogate model of the objective, and using this surrogate to reduce the number of computations of the objective
function in various ways.

Many surrogate modelling approaches have been used within 
Evolution Strategies (ESs) and Covariance Matrix Adaptation Evolution Strategy (CMA-ES): 
Radial Basis Functions network \cite{Hoffmann2006IEEE}, Gaussian Processes \cite{Ulmer2003CEC}, Artificial
Neural Network \cite{YJin2005}, Support Vector Regression \cite{KramerInformatica2010}, 
Local-Weighted Regression \cite{kernHansenMetaPPSN06,augerEvoNum2010},
Ranking Support Vector Machine (Ranking SVM) \cite{runarssonPPSN06,rankSurrogatePPSN10,Runarsson2011ISDA}.
In most cases, the surrogate model is used as a filter (to select $\lambda_{Pre}$ promising pre-children) and/or to estimate the fitness of some individuals in the current population.
An example of surrogate-assisted CMA-ES with filtering strategy can be found in \cite{rankSurrogatePPSN10}.

A well-known drawback of surrogate-assisted optimization is a strong dependence of the results on hyper-parameters used to build the surrogate model. 
Some optimal settings of hyper-parameters for a specific set of problems can be found by offline tuning, however for a new problem they are unknown in the black-box scenario. Moreover, the optimal hyper-parameters may dynamically change during the optimization of the function. 

Motivated by this open issues, new self-adapted surrogate-assisted \saACM\ algorithm have been proposed combining surrogate-assisted optimization of the expensive function and online optimization of the surrogate model hyper-parameters \cite{ACMGECCO2012}.

\section{The Algorithms}

\subsection{The \mulCMA }

In each iteration $t$, \mulCMA\ \cite{HansenECJ01} samples $\lambda$ new solutions $x_i \in \Rd$, where $i=1,\ldots,\lambda$, and selects the best $\mu$ among them. 
These $\mu$ points update the distribution of parameters of the algorithm to increase the probability of successful steps in iteration $t+1$.
The sampling is defined by a multi-variate normal distribution, $\mathcal N(\textbf{$m$}^t,{\sigma^t}^2C^t)$, with current mean of distribution $\textbf{$m$}^t$, $D\times D$ covariance matrix $C^t$ and step-size ${\sigma^t}$.

The active version of the CMA-ES proposed in \cite{1830788,2006:JastrebskiArnold} introduces a weighted negative update of the covariance matrix taking into account the information about $\lambda-\mu$ worst points as well as about $\mu$ best ones. The new version improves CMA-ES on 9 out of 12 tested unimodal functions by a factor up to 2, and the advantages are more pronounced in larger dimension. While the new update scheme does not guarantee the positive-definiteness of the covariance matrix, it can be numerically controlled \cite{1830788}. 
Since in our study we do not observe any negative effects of this issue, we will use aCMA-ES, the active version of the CMA-ES, for comparison with the surrogate-assisted algorithms.

\subsection{The \sACM }

The \sACM\ \cite{ACMGECCO2012} is the surrogate-assisted version of the \mulCMA, where the surrogate model is used periodically instead of the expensive function for direct optimization.
The use of Ranking SVM allows to preserve the property of CMA-ES of invariance with respect to rank-preserving transformation of the fitness function. The property of invariance with respect to the orthogonal transformation of the search space is preserved thanks to the definition of the kernel function by the covariance matrix, adapted during the search. 

In \sACM\ we perform the following surrogate-assisted optimization loop: we optimize the surrogate model $\hat{f}$ for $\hat{n}$ generations by the CMA-ES, then we continue and optimize the expensive function $f(x)$ for one generation. To adjust the number of generations $\hat{n}$ for the next time, the model error can be computed as a fraction of incorrectly predicted comparison relations that we observe, when we compare the ranking of the last $\lambda$ evaluated points according to $f(x)$ and $\hat{f}$. The \sACM\ uses the generation of the CMA-ES as a black-box procedure, and it has been shown in \cite{ACMGECCO2012}, that the improvement of the CMA-ES from passive to active version (aCMA-ES) leads to a comparable improvement of its surrogate-assisted versions (\sACM\ and \saACM).

The main novelty of the \sACM\ is the online optimization of the surrogate model hyper-parameters during the optimization of the fitness function. The algorithm performs the search in the space of model hyper-parameters, generating $\lambda_{hyp}$ surrogate models in each iteration. The fitness of the model can be measured as a prediction error of the ranking on the last $\lambda$ evaluated points. This allows the user to define only the range of hyper-parameters and let algorithm to find the most suitable values for the current iteration $t$. 

The detailed description of \sACM\ is given in \cite{ACMGECCO2012}.

\subsection{The Benchmarked Algorithms }

For benchmarking we consider four CMA-ES algorithms in restart scenario: IPOP-aCMA-ES \cite{1830788}, BIPOP-CMA-ES \cite{Hansen:2009:BIPOP}, \IPOPsaACM\ and \BIPOPsaACM \cite{ACMGECCO2012}. 
For \IPOPsaACM\ and \BIPOPsaACM\ we use the same parameters of the CMA-ES and termination criteria in IPOP and BIPOP scenario as in the original papers. 
The default parameters for \sACM\ algorithms are given in \cite{ACMGECCO2012}.

%
\section{Results}

\begin{figure*}
\centering
\begin{tabular}{@{}c@{}c@{}c@{}c@{}}
\includegraphics[width=0.253\textwidth, trim= 0.7cm 0.8cm 0.5cm 0.5cm, clip]{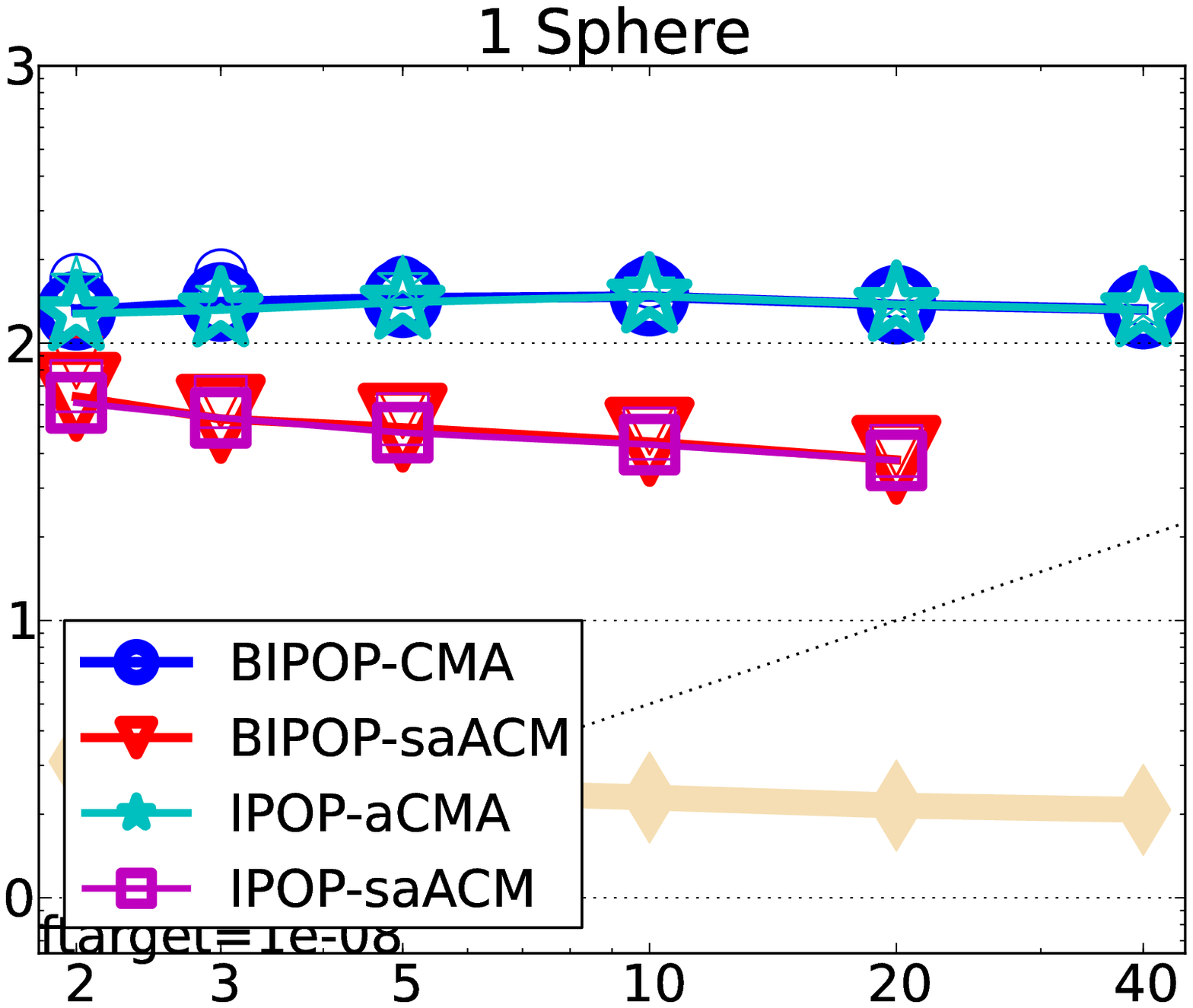}&
\includegraphics[width=0.238\textwidth, trim= 1.8cm 0.8cm 0.5cm 0.5cm, clip]{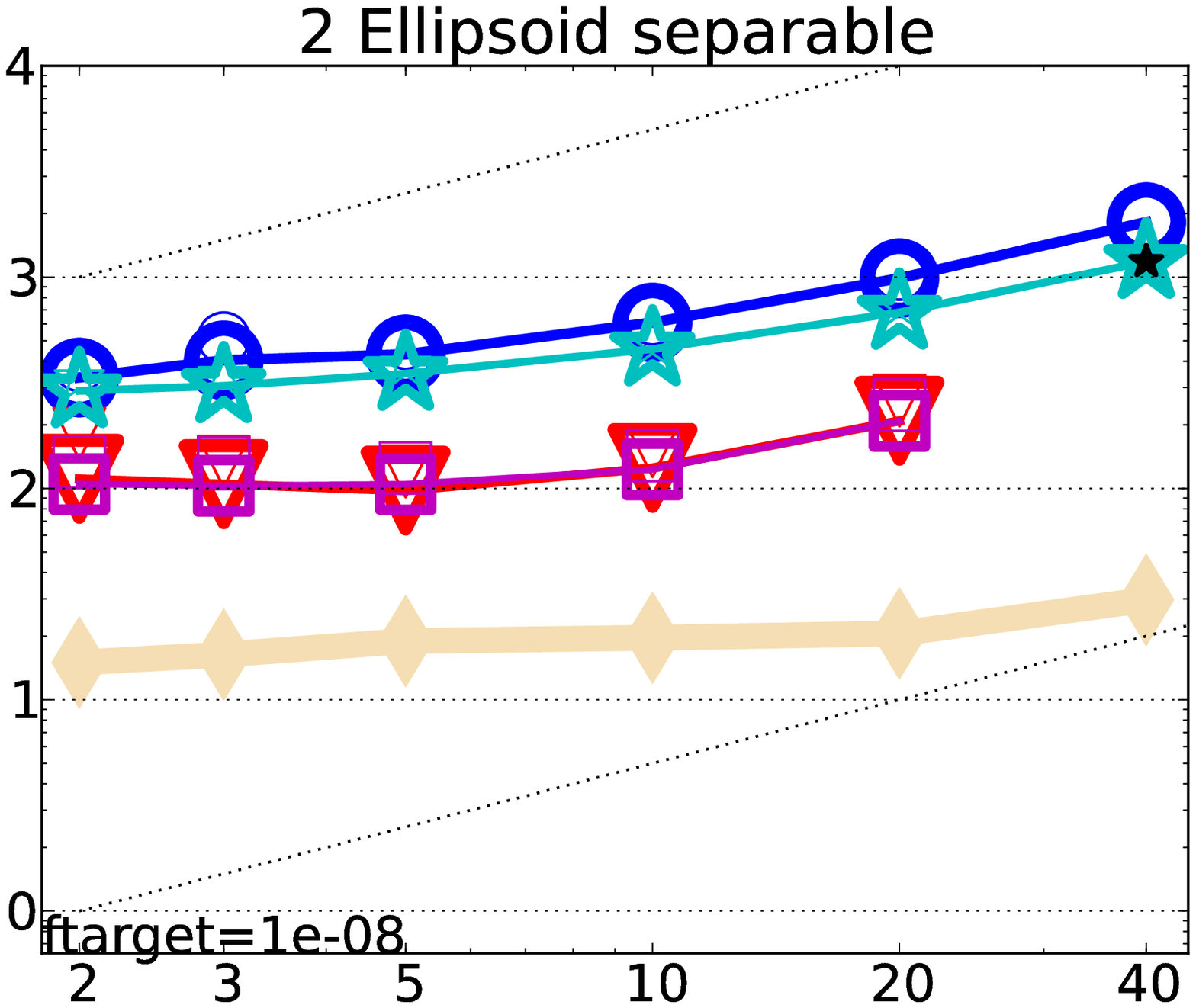}&
\includegraphics[width=0.238\textwidth, trim= 1.8cm 0.8cm 0.5cm 0.5cm, clip]{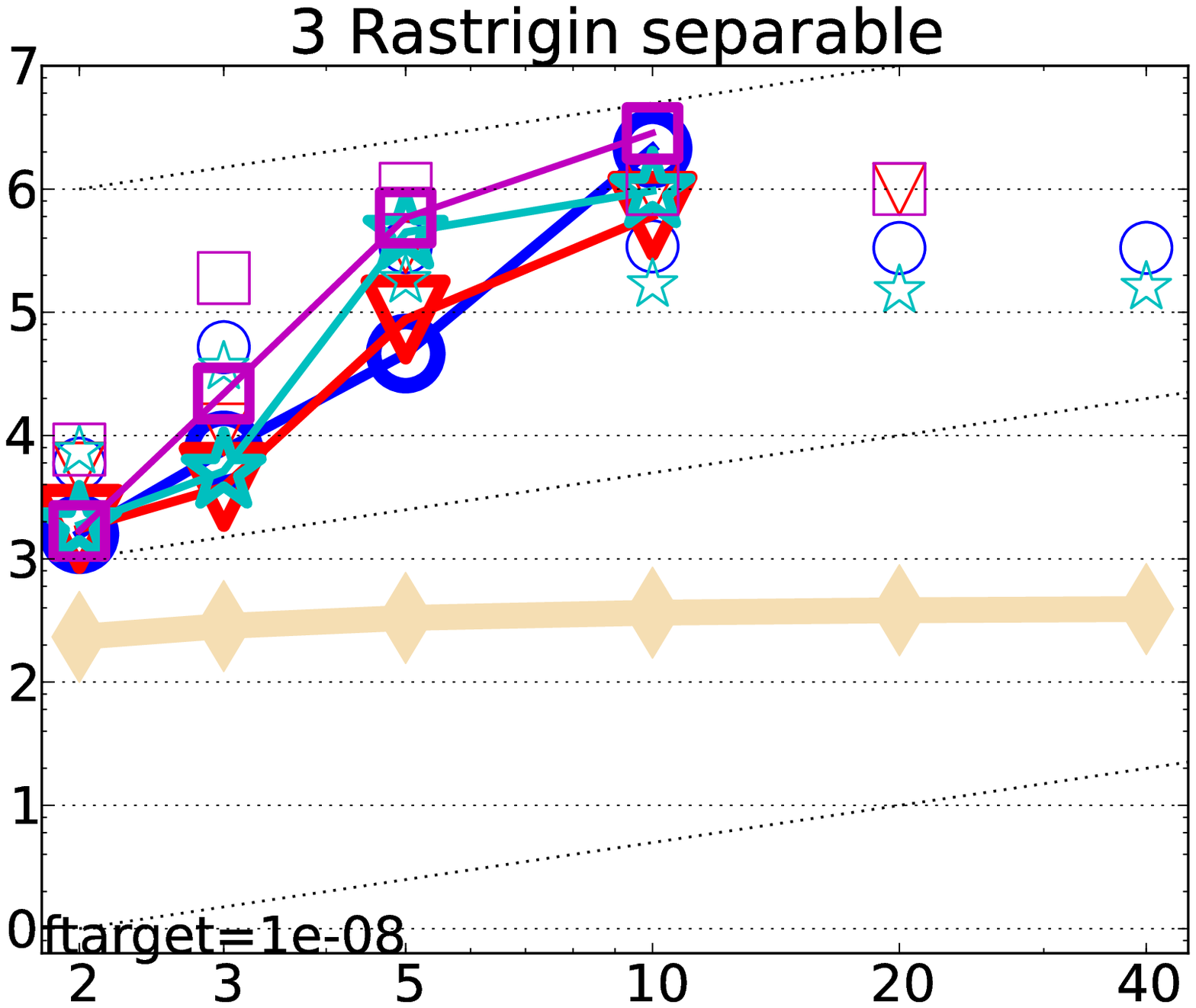}&
\includegraphics[width=0.238\textwidth, trim= 1.8cm 0.8cm 0.5cm 0.5cm, clip]{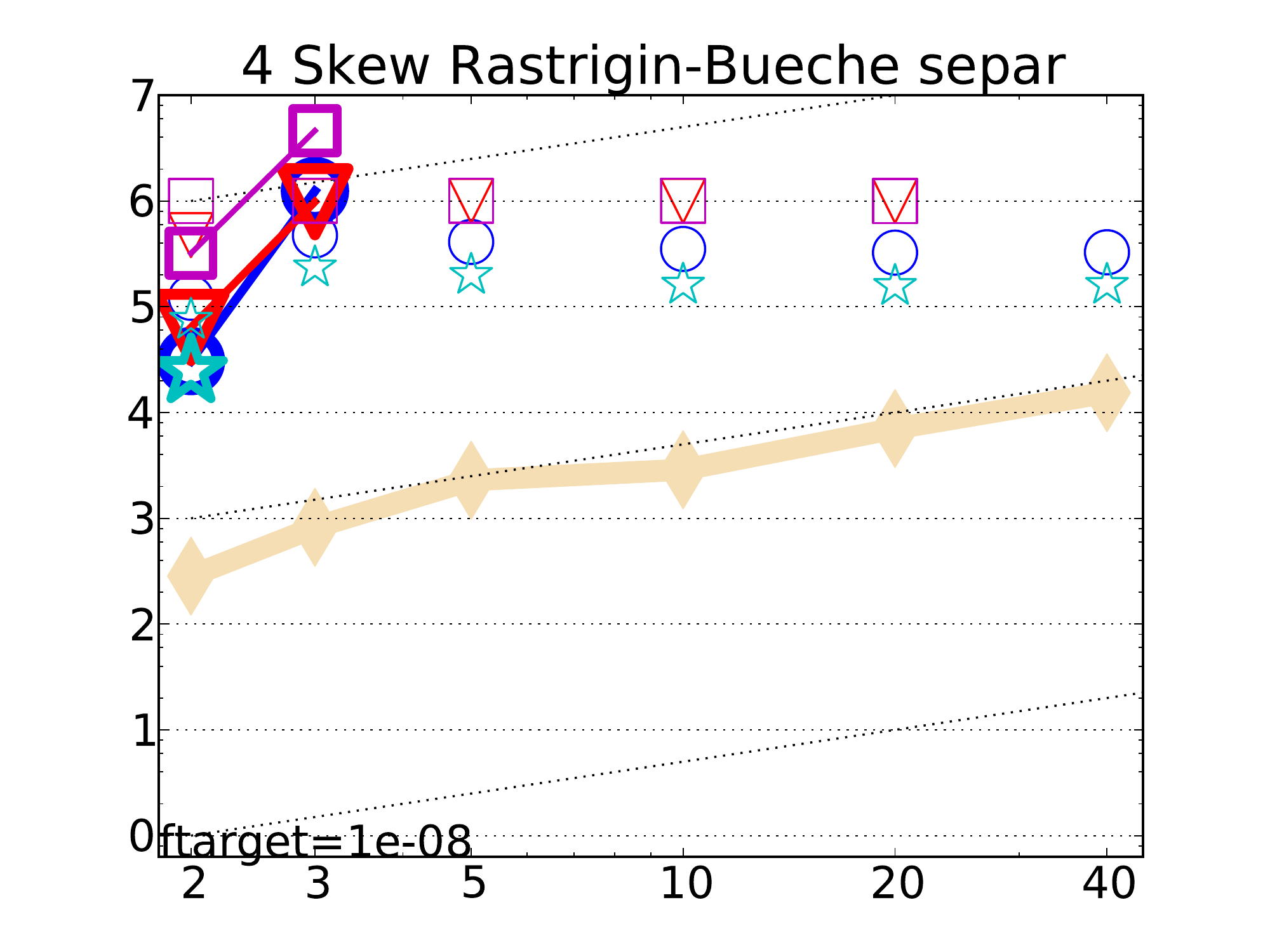}\\
\includegraphics[width=0.253\textwidth, trim= 0.7cm 0.8cm 0.5cm 0.5cm, clip]{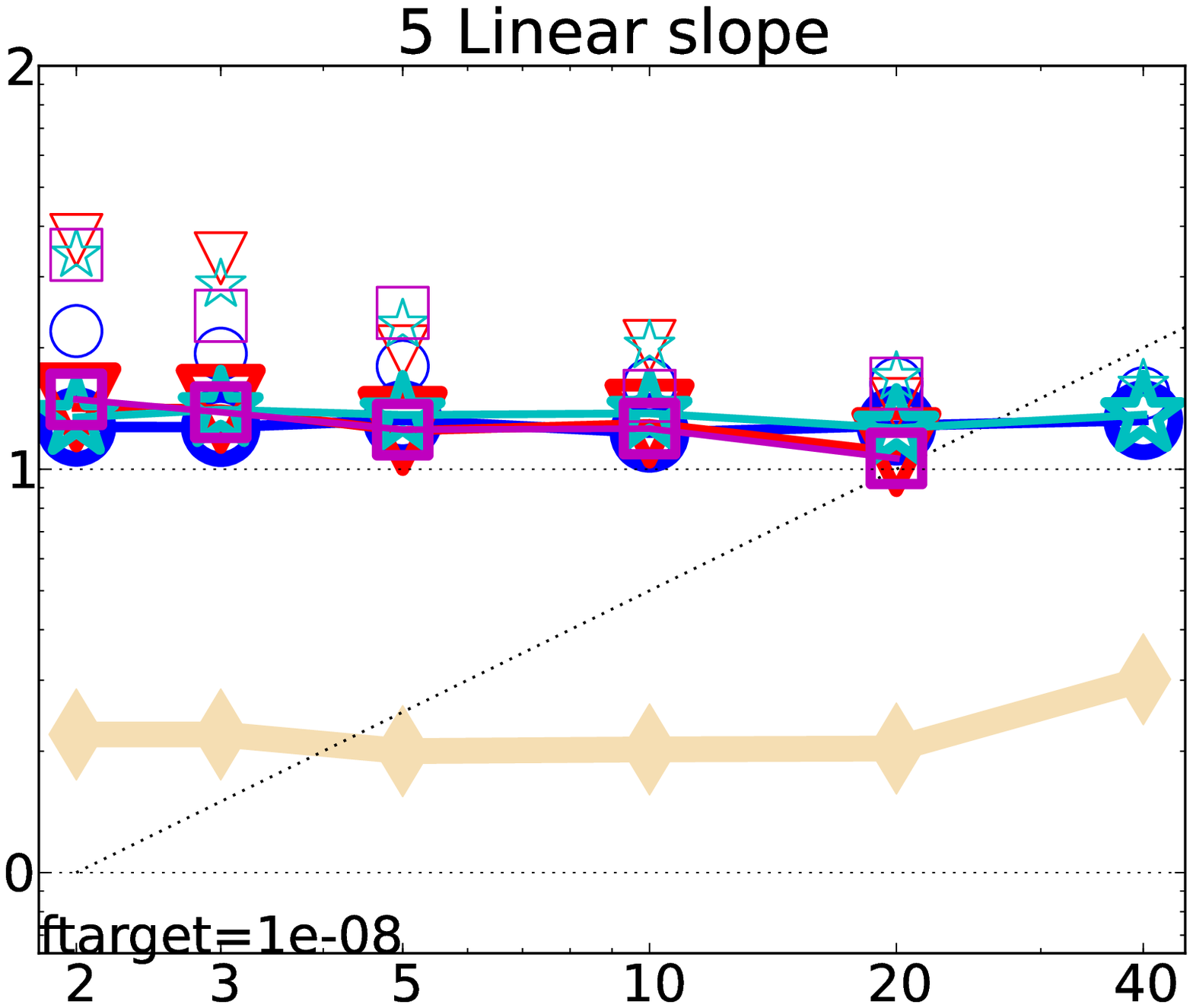}&
\includegraphics[width=0.238\textwidth, trim= 1.8cm 0.8cm 0.5cm 0.5cm, clip]{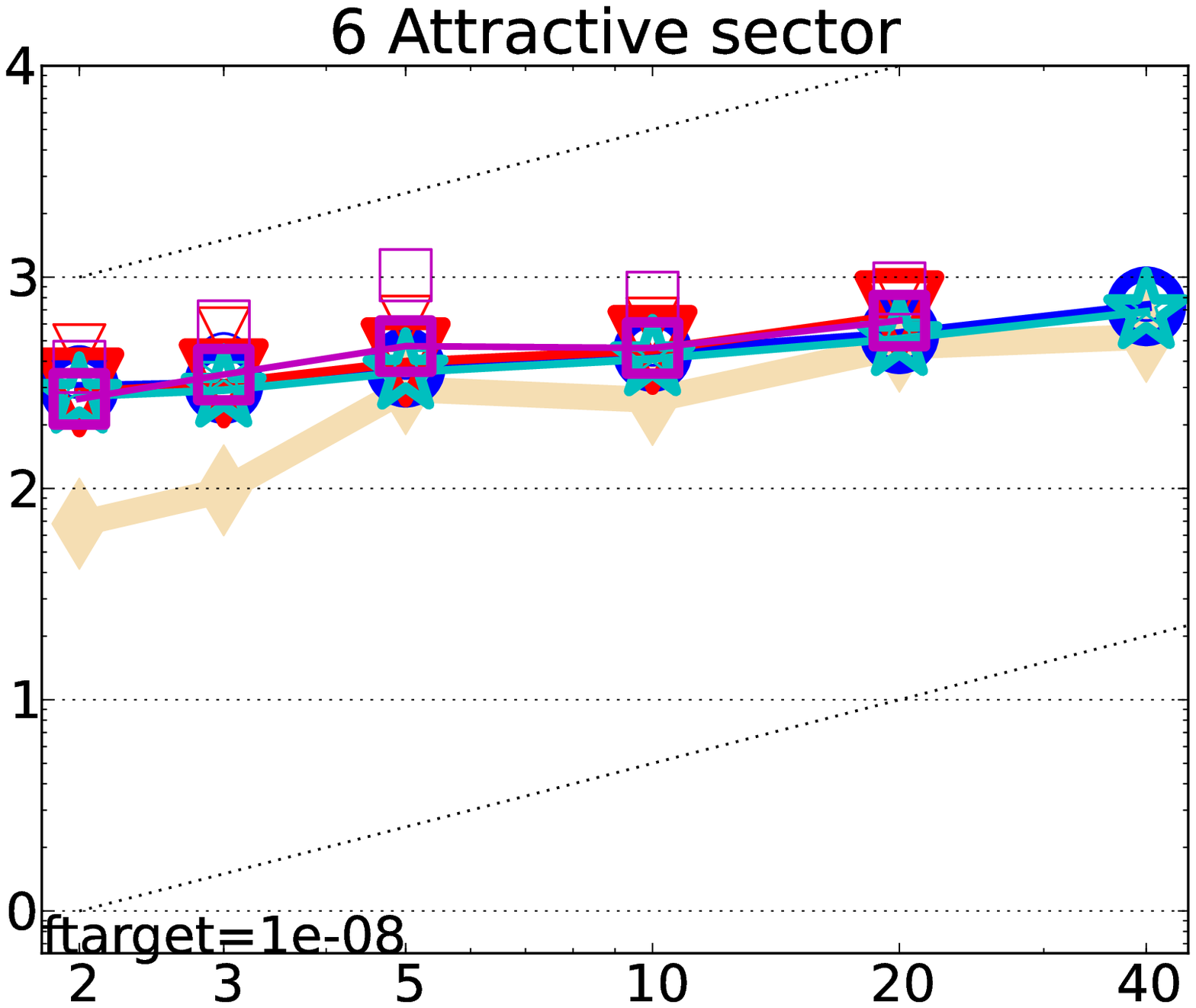}&
\includegraphics[width=0.238\textwidth, trim= 1.8cm 0.8cm 0.5cm 0.5cm, clip]{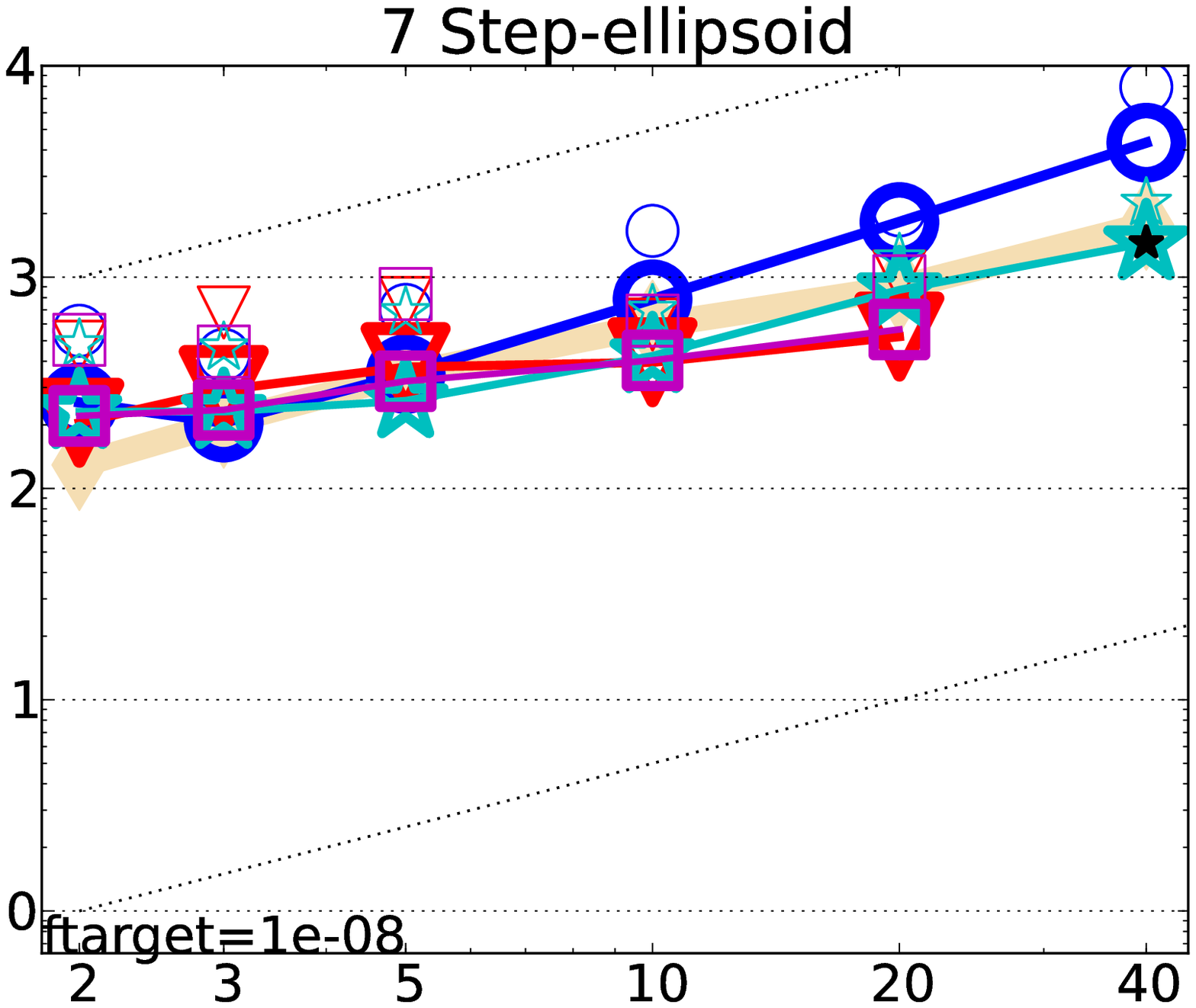}&
\includegraphics[width=0.238\textwidth, trim= 1.8cm 0.8cm 0.5cm 0.5cm, clip]{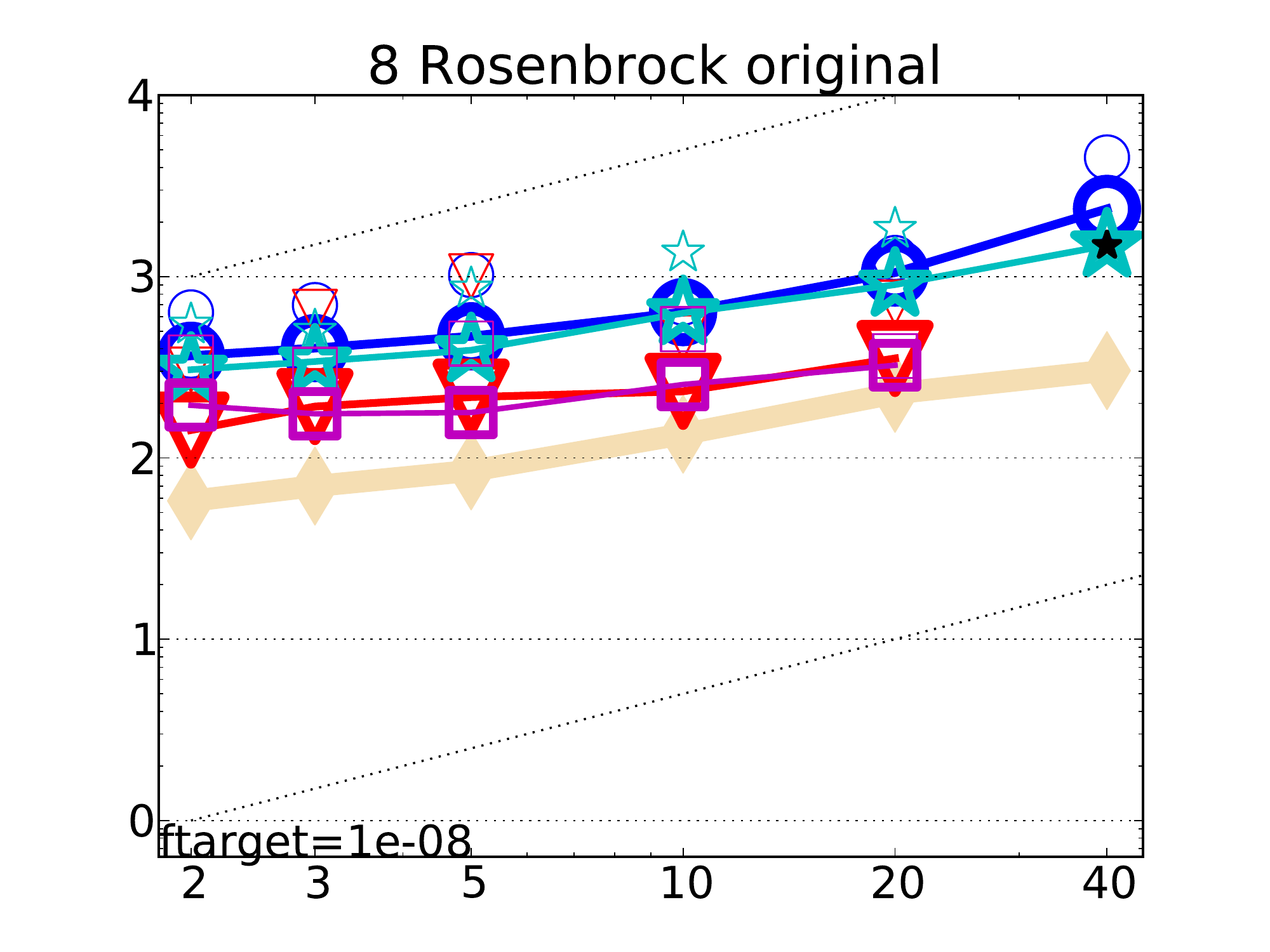}\\
\includegraphics[width=0.253\textwidth, trim= 0.7cm 0.8cm 0.5cm 0.5cm, clip]{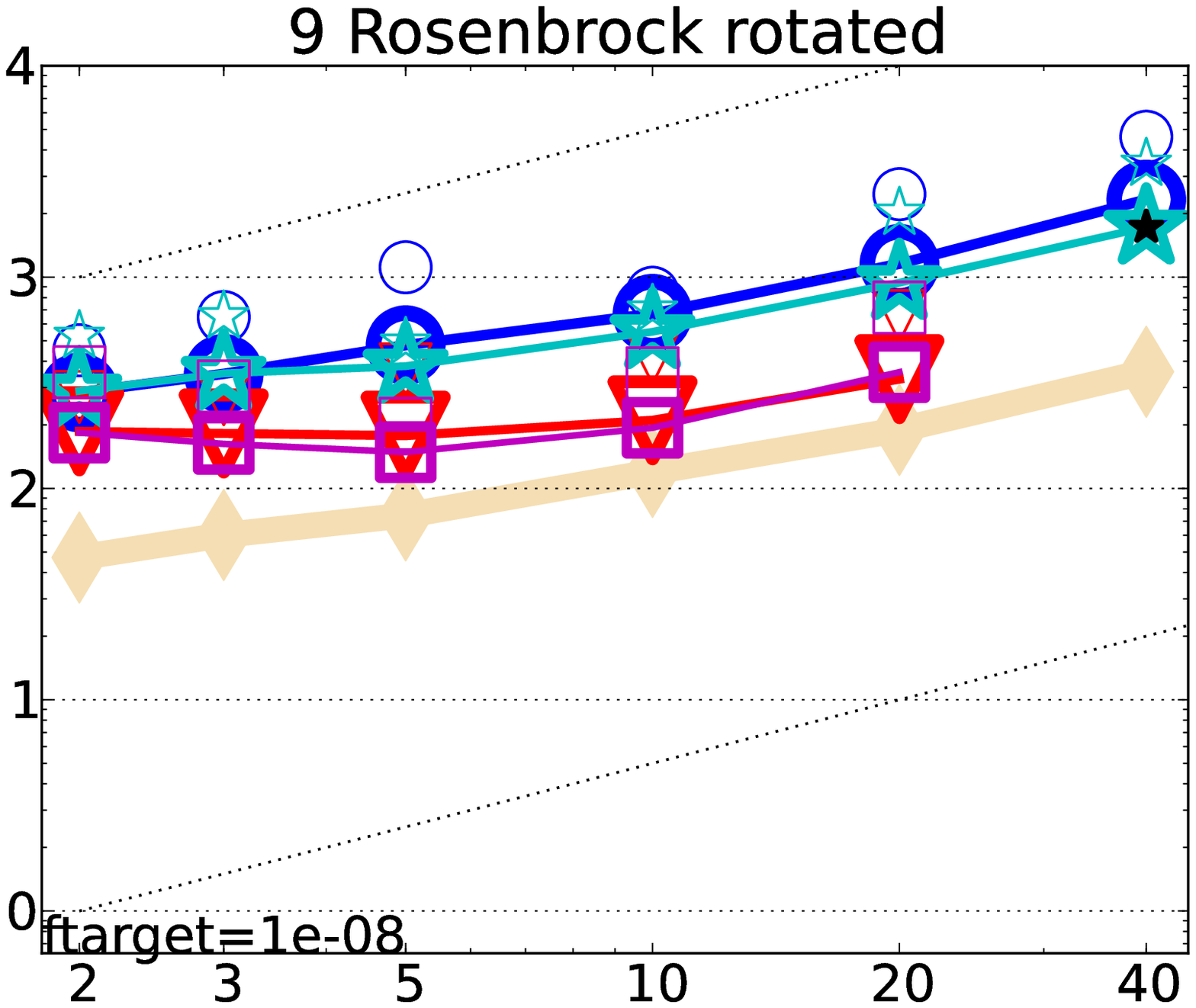}&
\includegraphics[width=0.238\textwidth, trim= 1.8cm 0.8cm 0.5cm 0.5cm, clip]{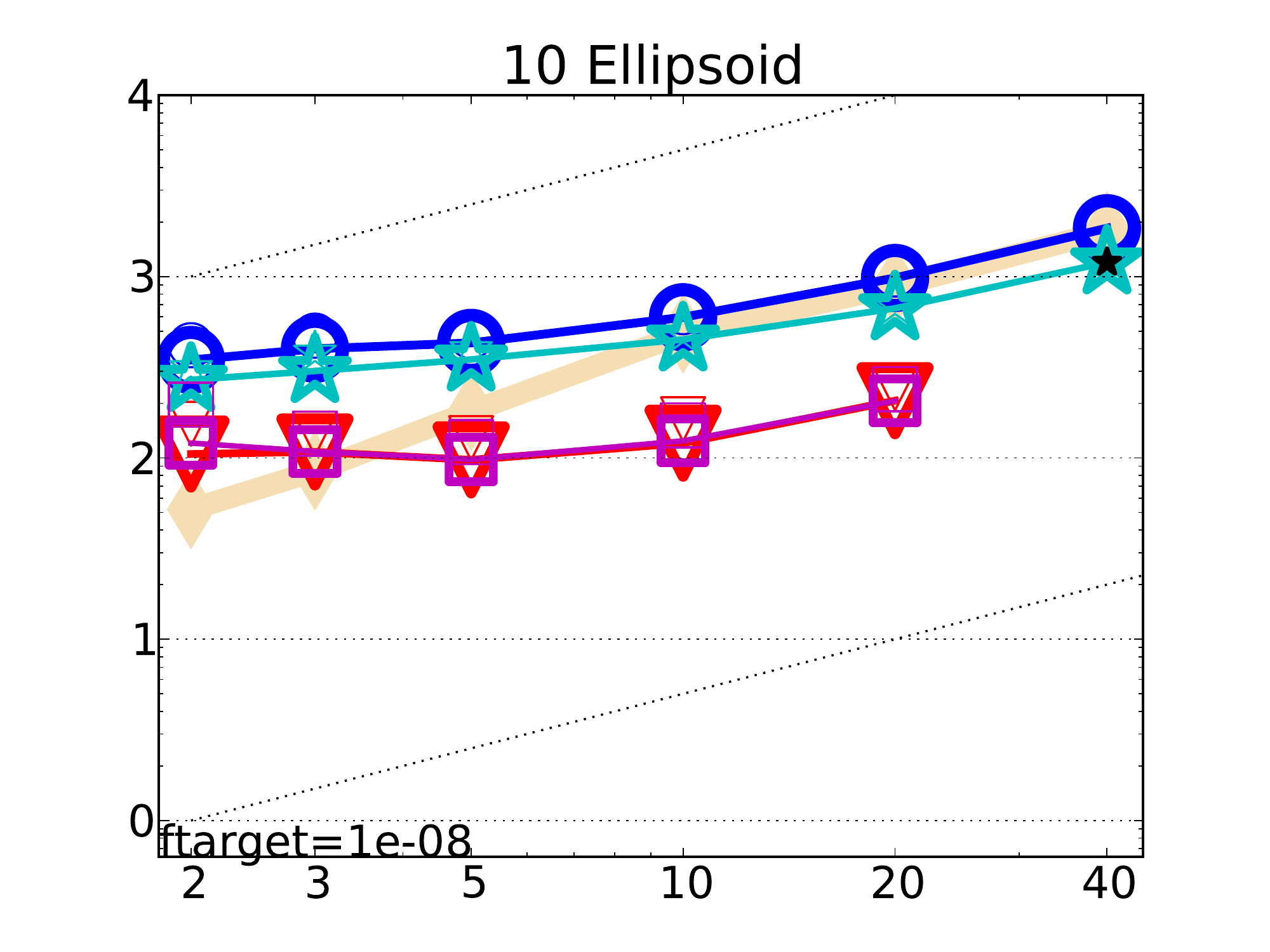}&
\includegraphics[width=0.238\textwidth, trim= 1.8cm 0.8cm 0.5cm 0.5cm, clip]{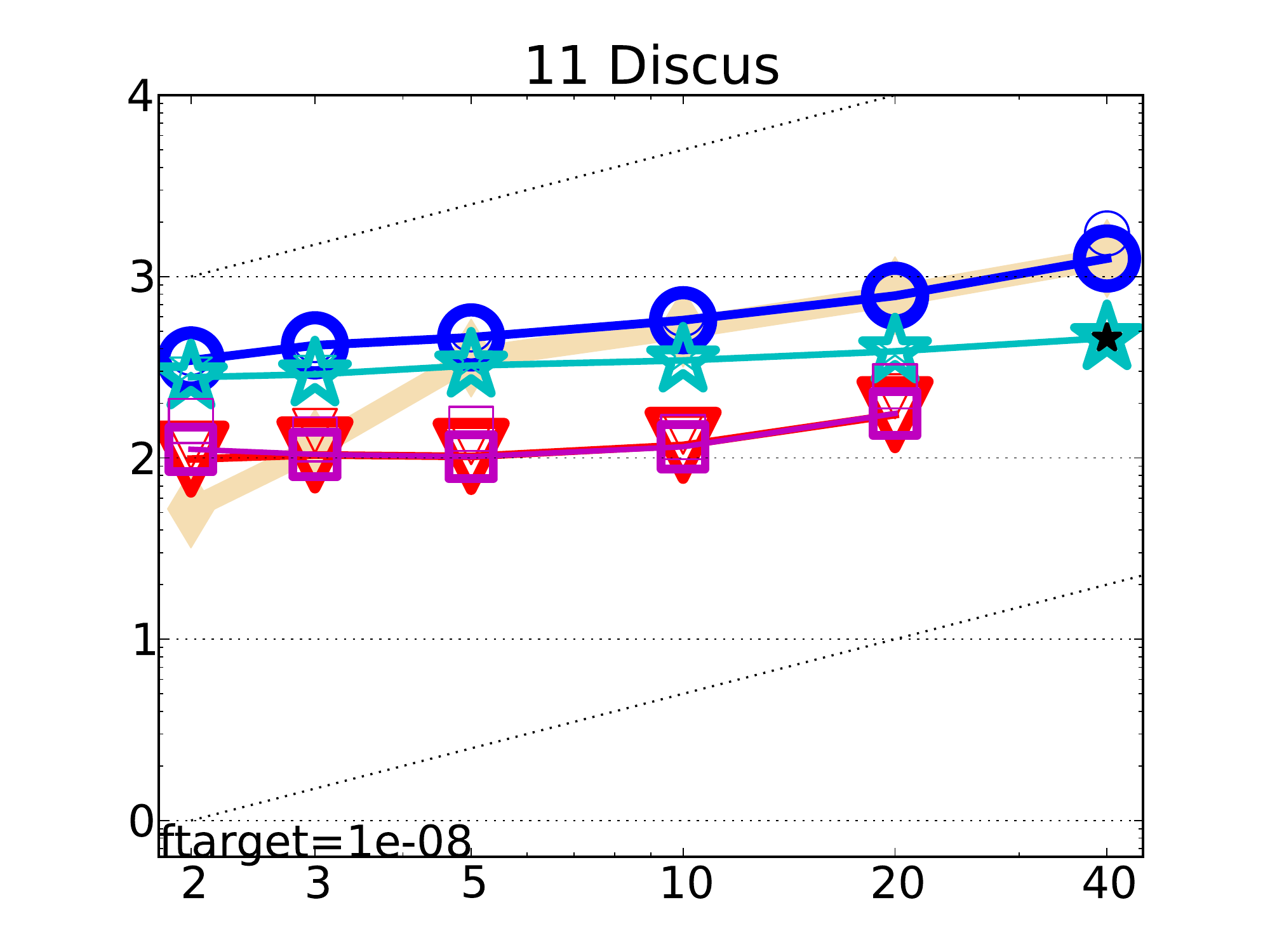}&
\includegraphics[width=0.238\textwidth, trim= 1.8cm 0.8cm 0.5cm 0.5cm, clip]{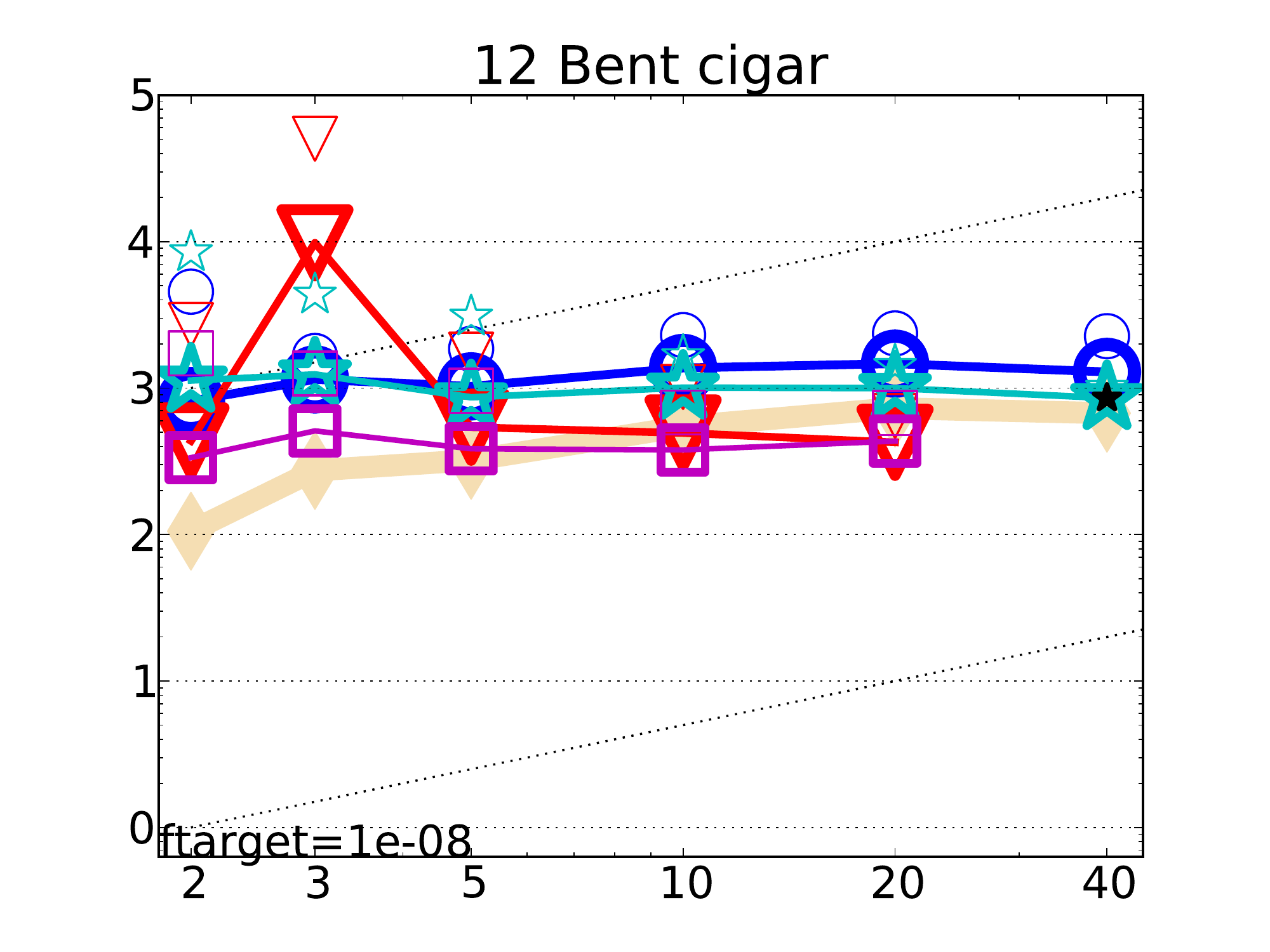}\\
\includegraphics[width=0.253\textwidth, trim= 0.7cm 0.8cm 0.5cm 0.5cm, clip]{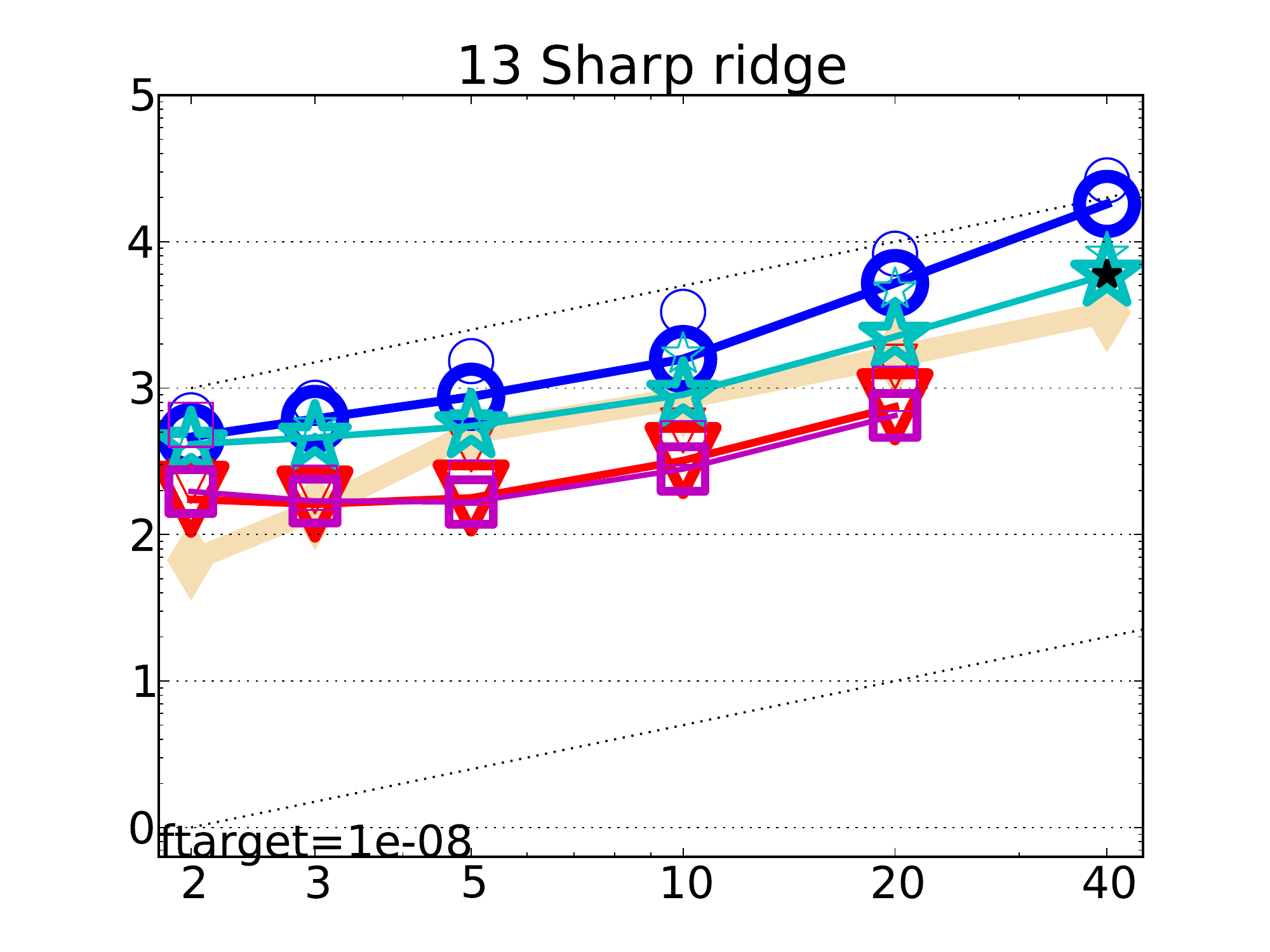}&
\includegraphics[width=0.238\textwidth, trim= 1.8cm 0.8cm 0.5cm 0.5cm, clip]{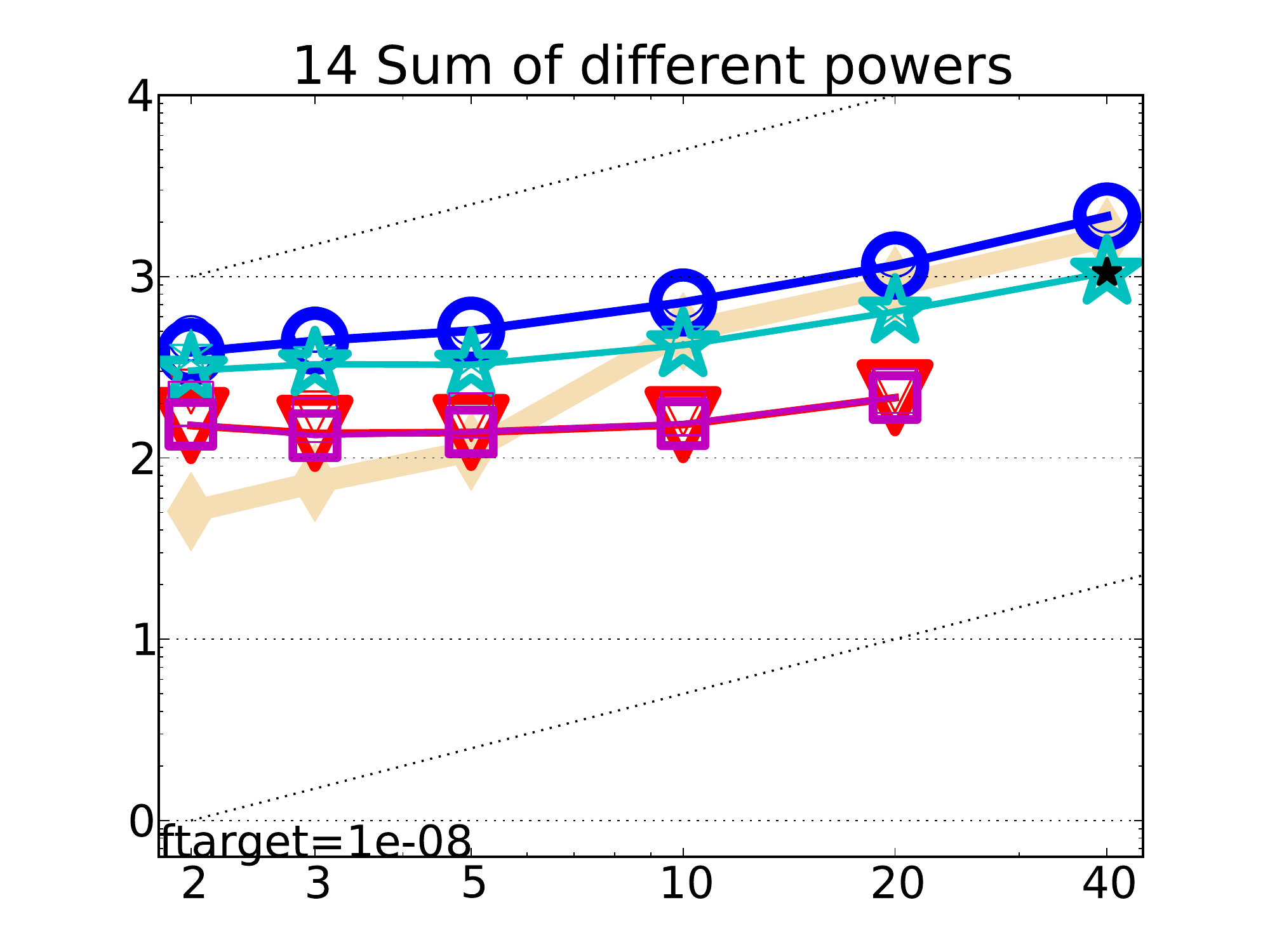}&
\includegraphics[width=0.238\textwidth, trim= 1.8cm 0.8cm 0.5cm 0.5cm, clip]{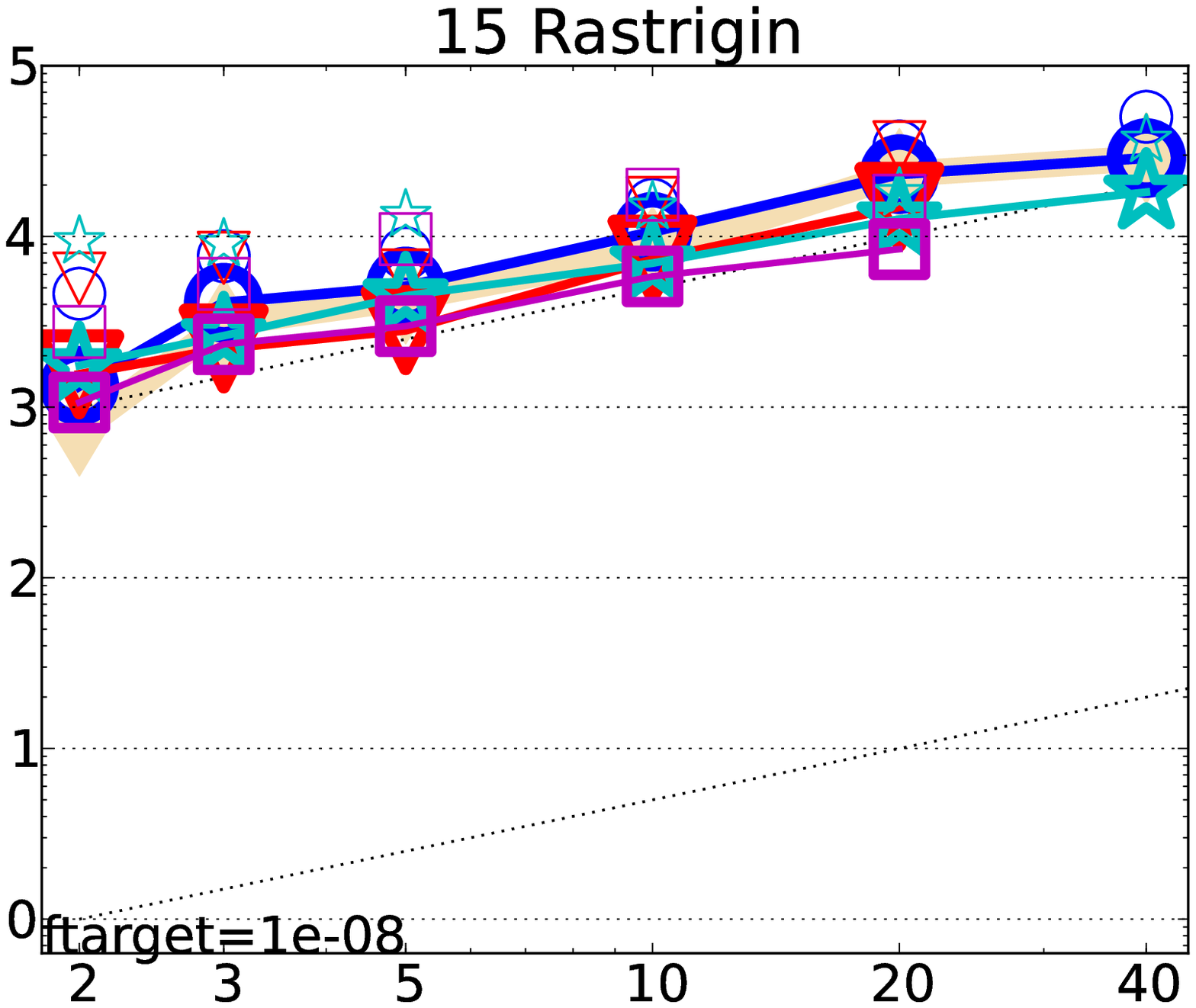}&
\includegraphics[width=0.238\textwidth, trim= 1.8cm 0.8cm 0.5cm 0.5cm, clip]{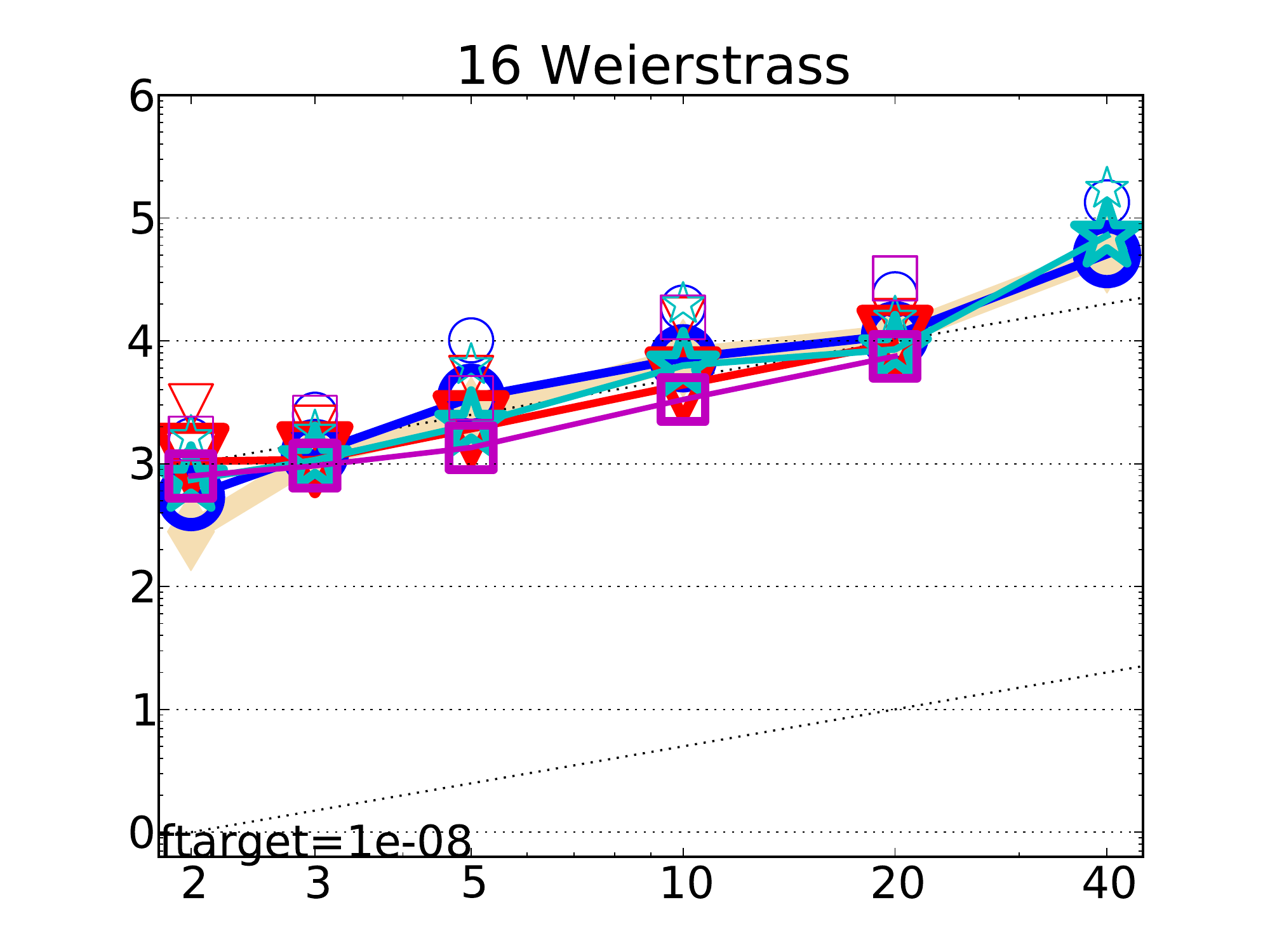}\\
\includegraphics[width=0.253\textwidth, trim= 0.7cm 0.8cm 0.5cm 0.5cm, clip]{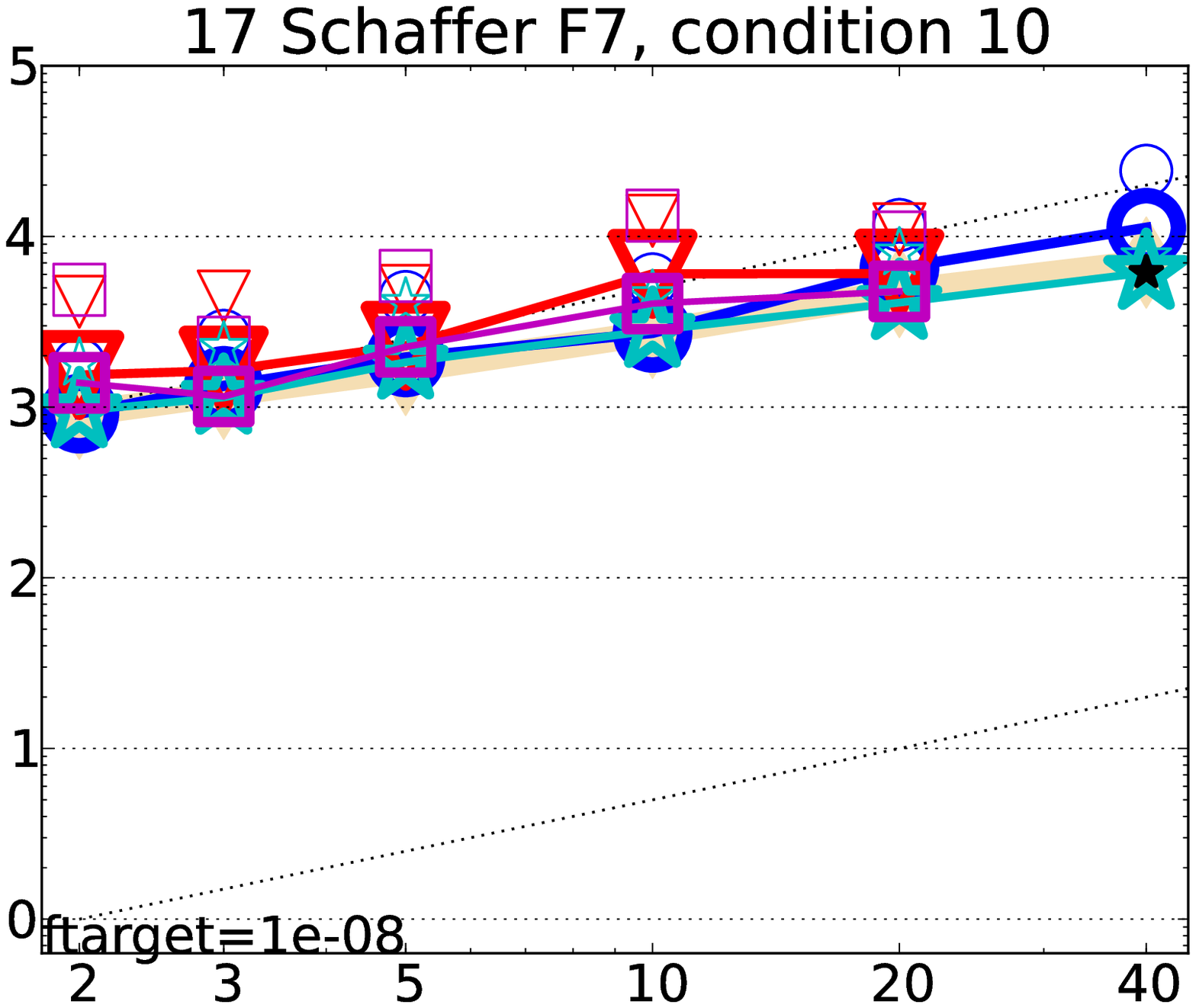}&
\includegraphics[width=0.238\textwidth, trim= 1.8cm 0.8cm 0.5cm 0.5cm, clip]{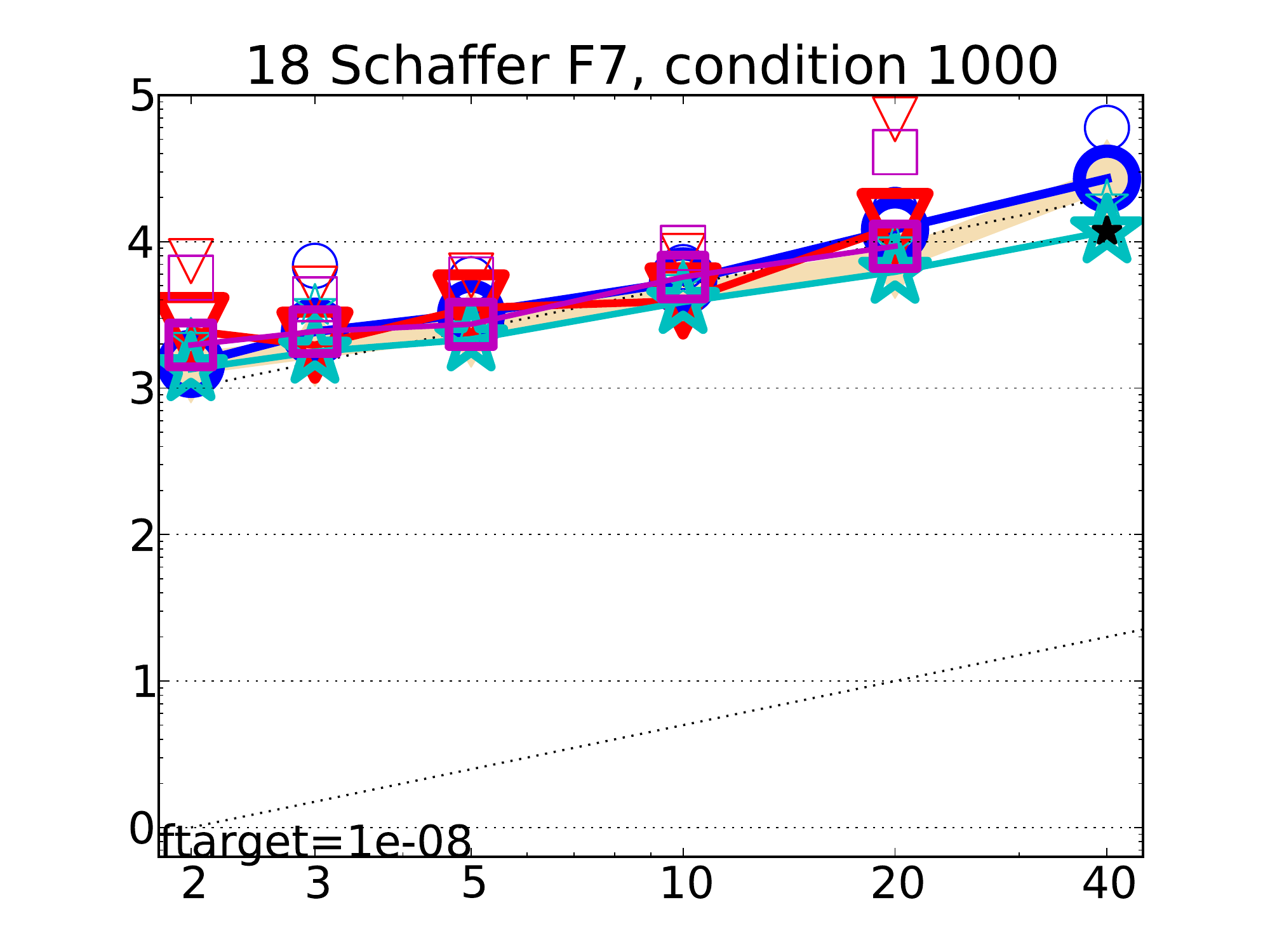}&
\includegraphics[width=0.238\textwidth, trim= 1.8cm 0.8cm 0.5cm 0.5cm, clip]{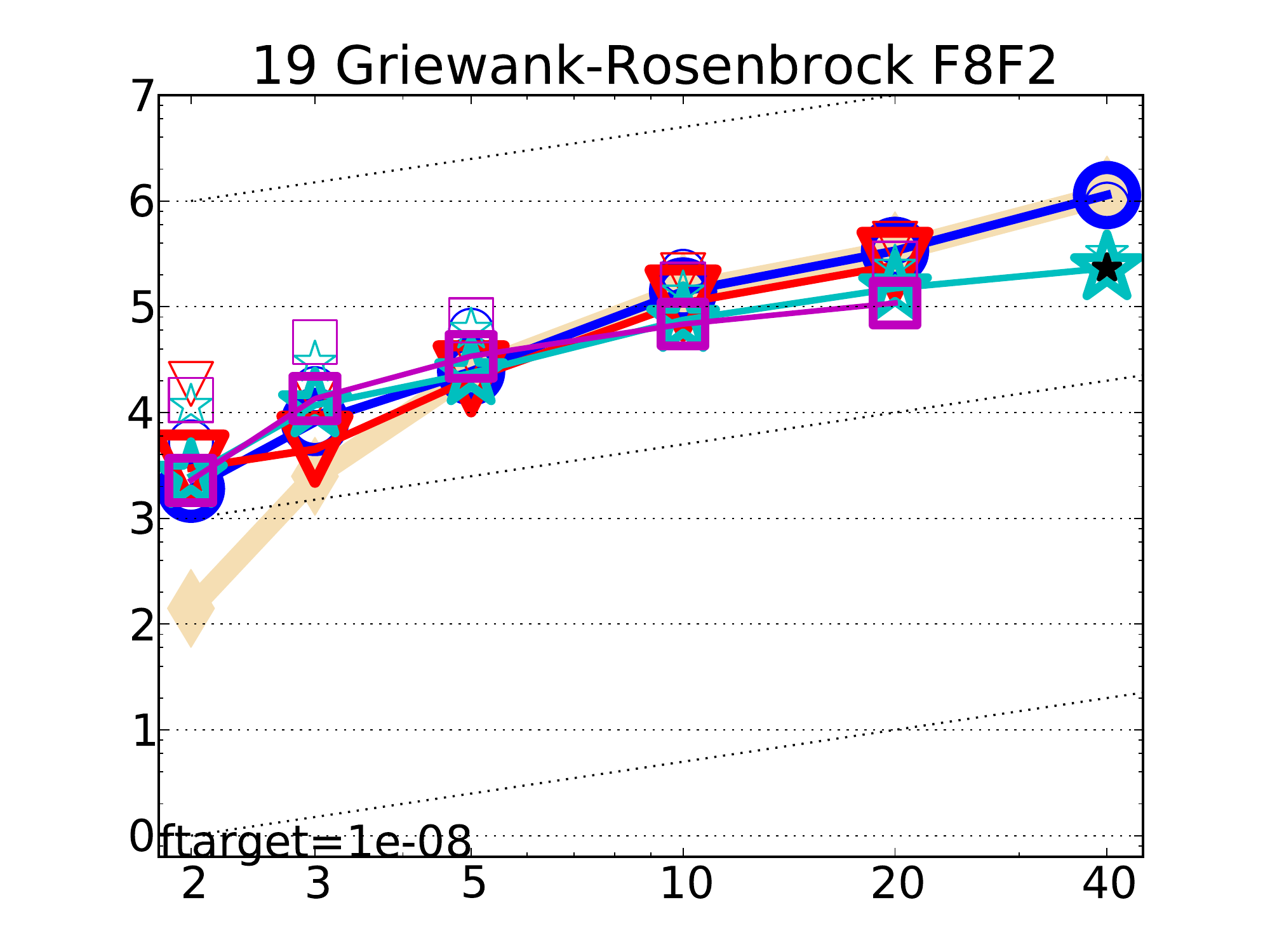}&
\includegraphics[width=0.238\textwidth, trim= 1.8cm 0.8cm 0.5cm 0.5cm, clip]{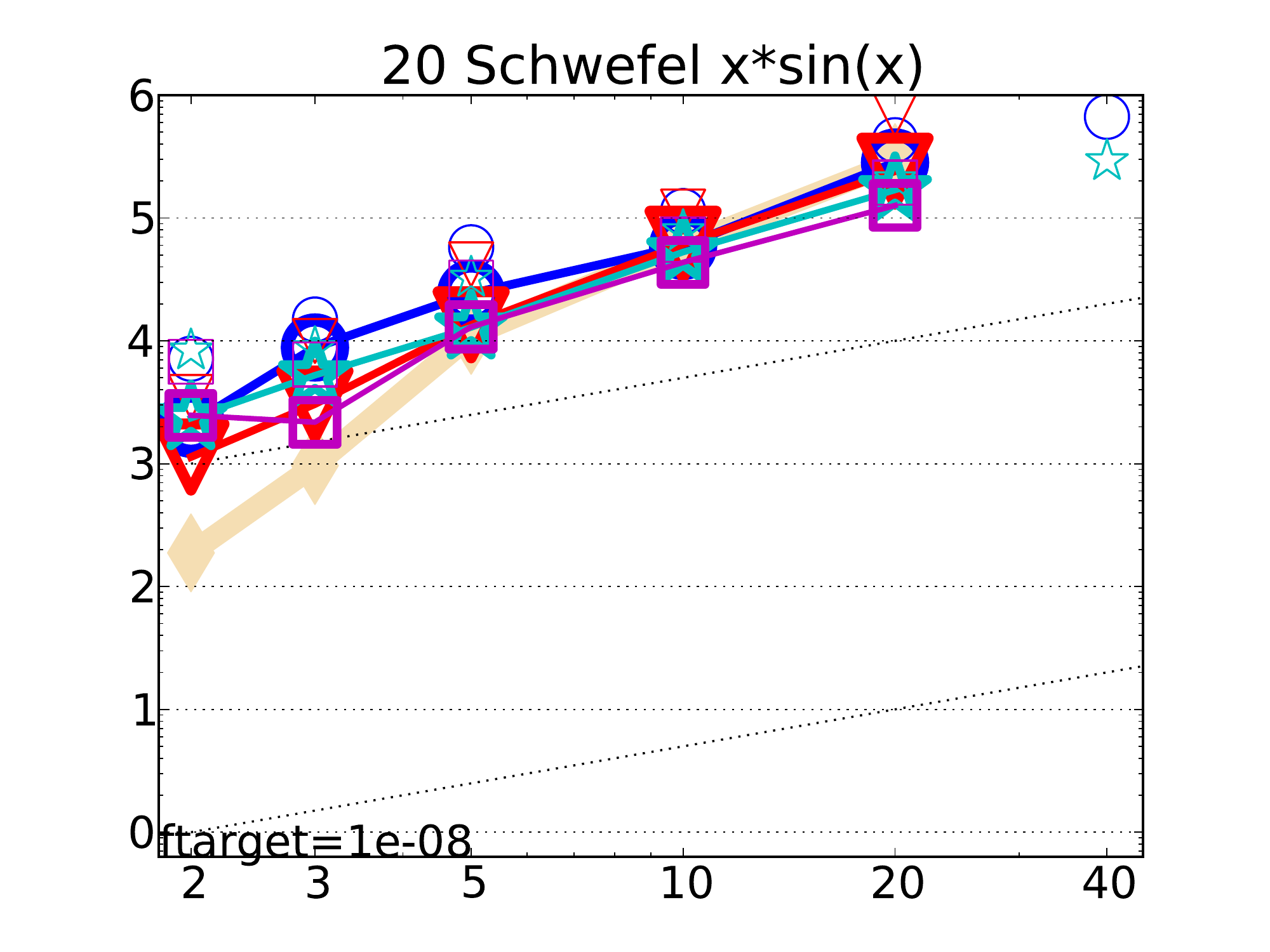}\\
\includegraphics[width=0.253\textwidth, trim= 0.7cm 0.0cm 0.5cm 0.5cm, clip]{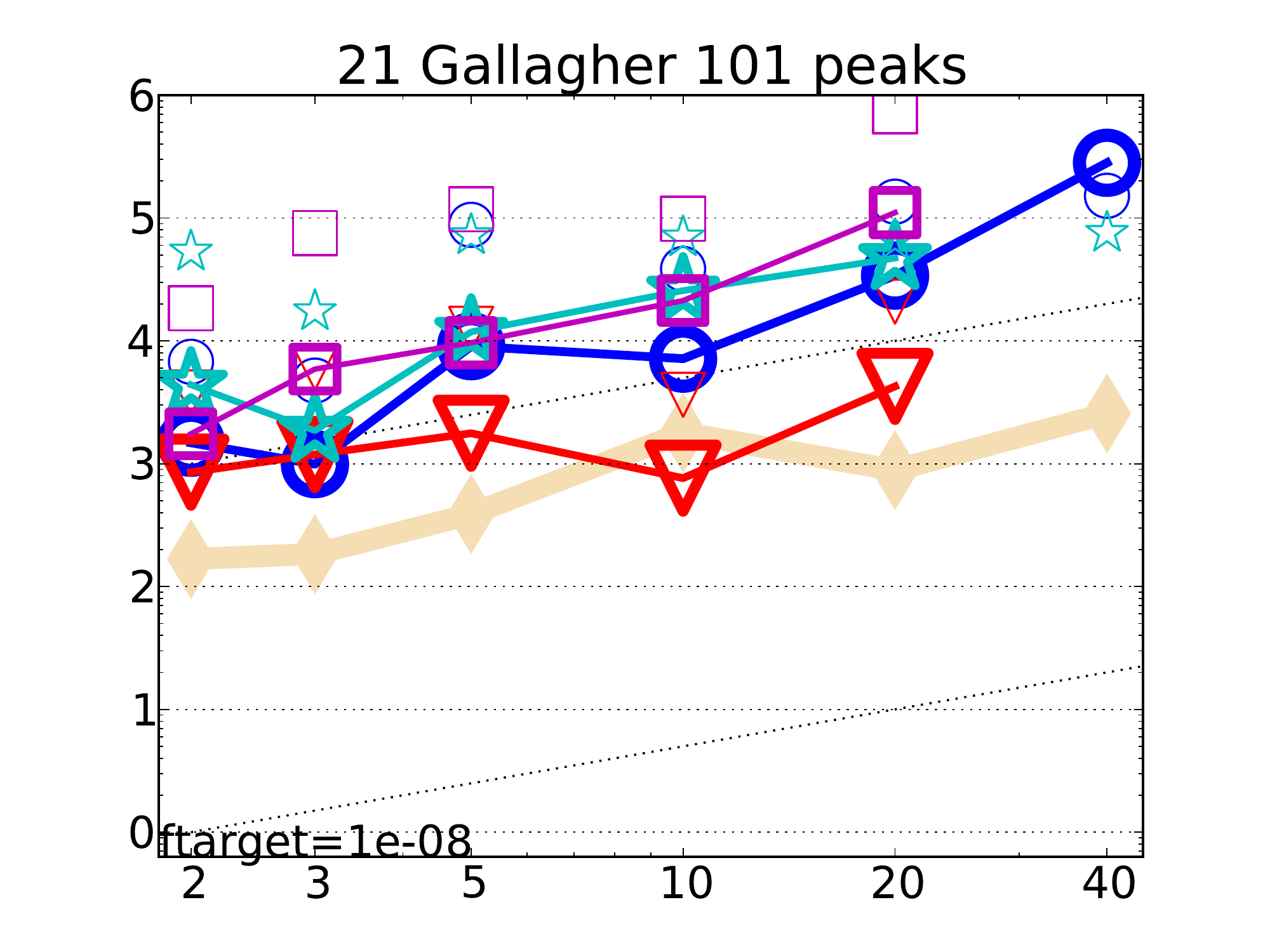}&
\includegraphics[width=0.238\textwidth, trim= 1.8cm 0.0cm 0.5cm 0.5cm, clip]{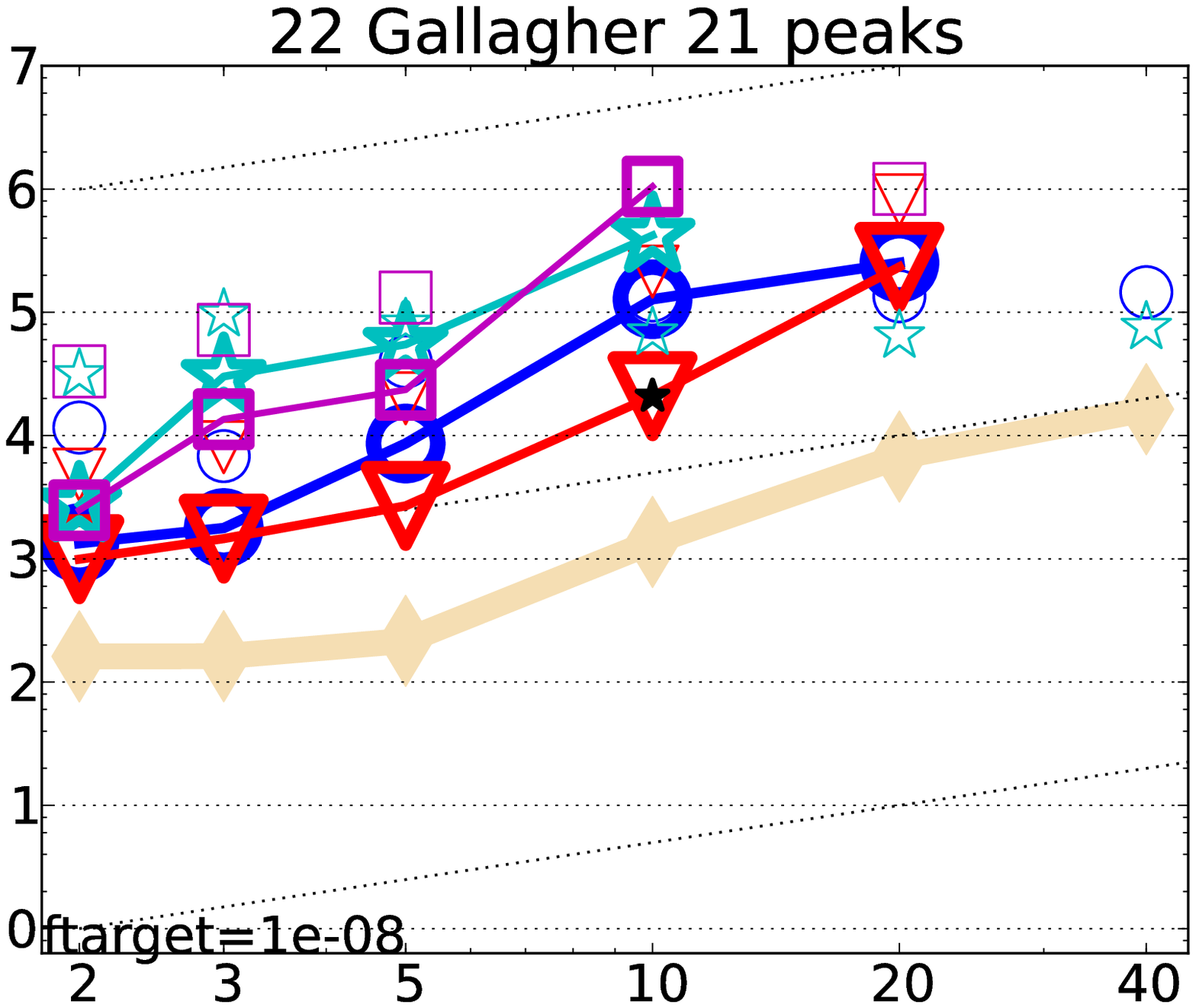}&
\includegraphics[width=0.238\textwidth, trim= 1.8cm 0.0cm 0.5cm 0.5cm, clip]{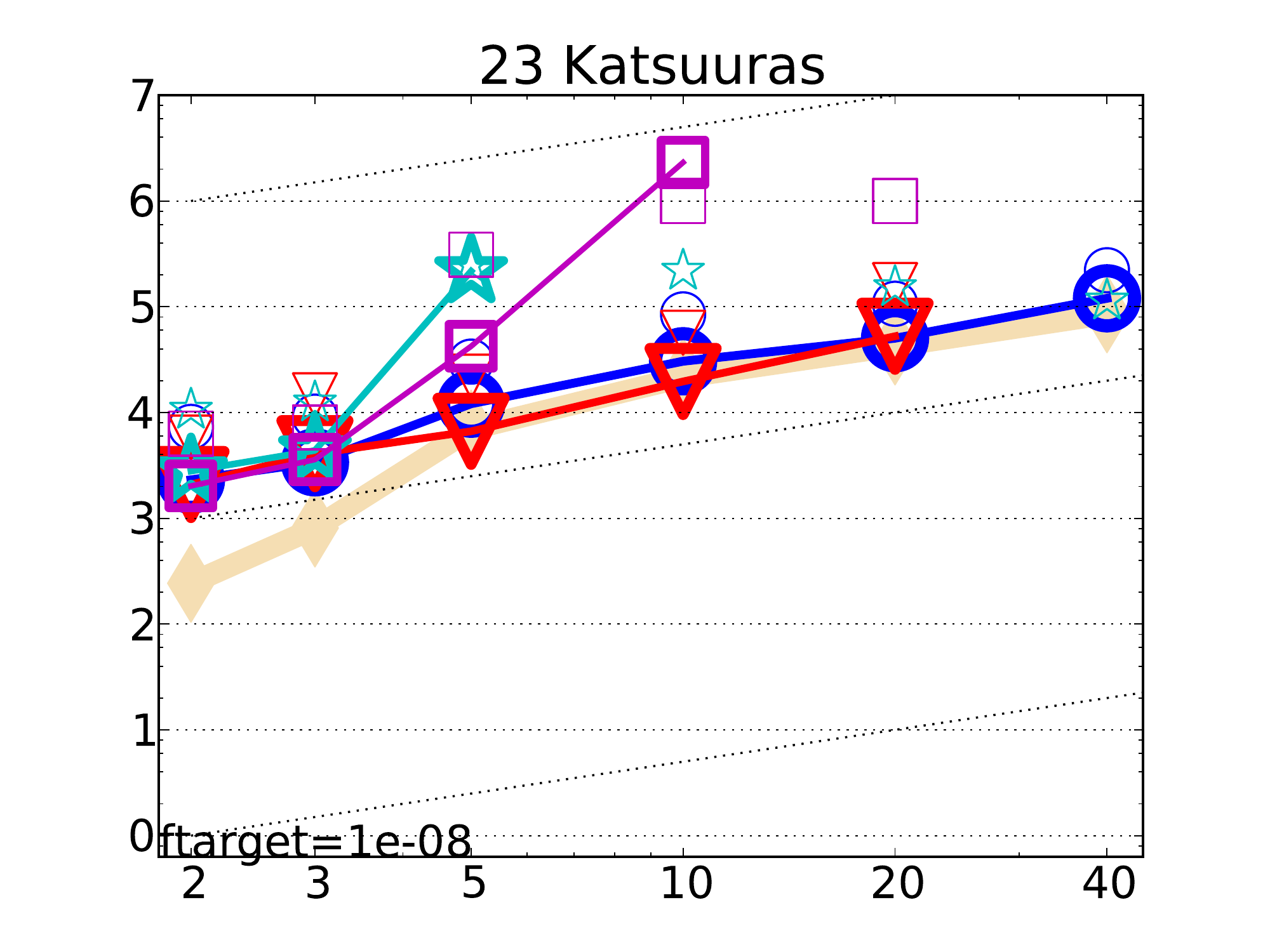}&
\includegraphics[width=0.238\textwidth, trim= 1.8cm 0.0cm 0.5cm 0.5cm, clip]{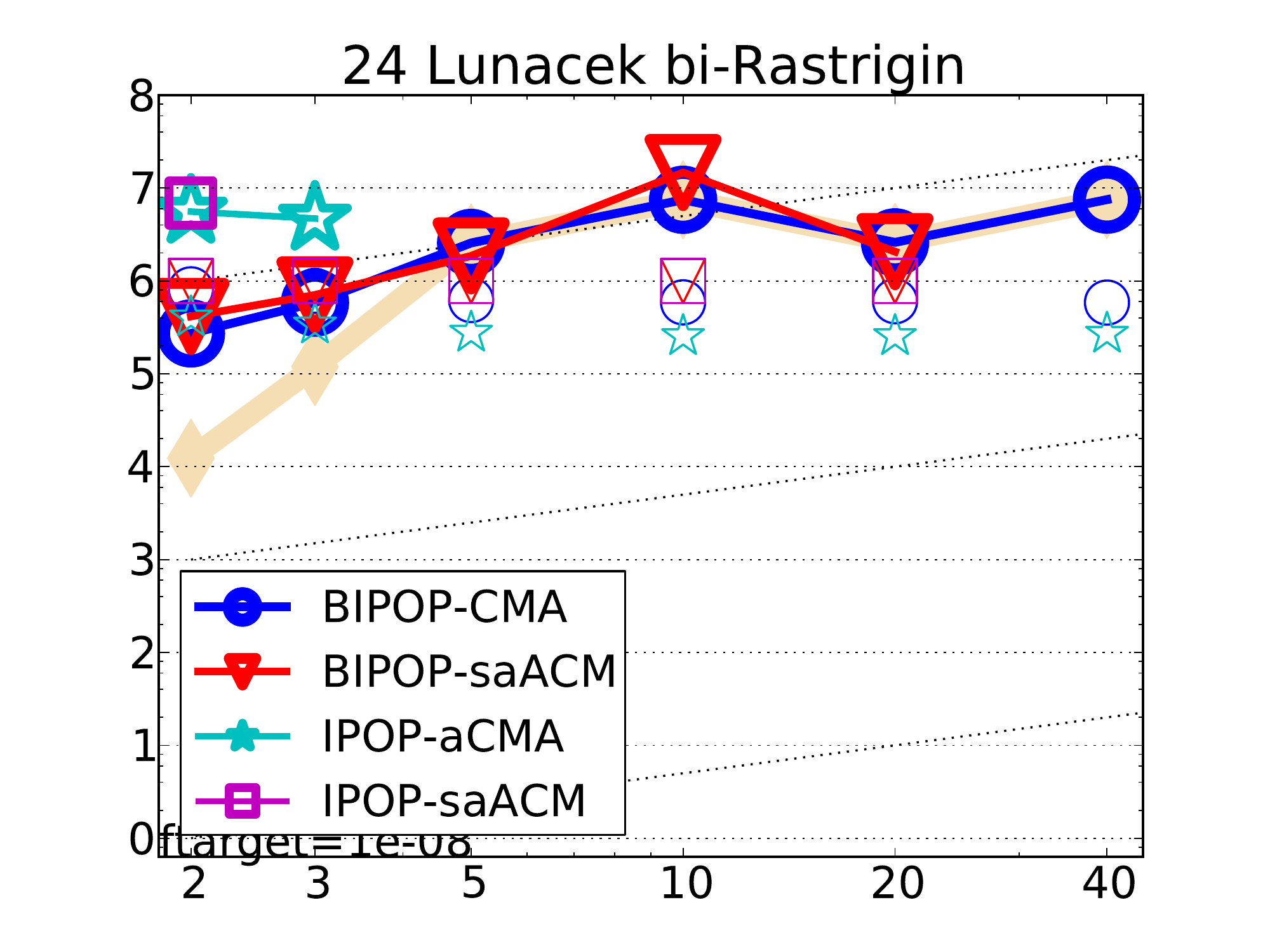}
\end{tabular}
\vspace*{-0.2cm}
\caption[Expected running time divided by dimension
versus dimension]{
\label{fig:scaling}
\bbobppfigslegend{$f_1$ and $f_{24}$}.  
}
\end{figure*}

\begin{figure*}
\begin{tabular}{@{}c@{}c@{}}
 separable fcts & moderate fcts \\
  \input{\bbobdatapath pprldmany_05D_separ}%
  \includegraphics[width=0.4135\textwidth,trim=0mm 0mm 34mm 10mm, clip]{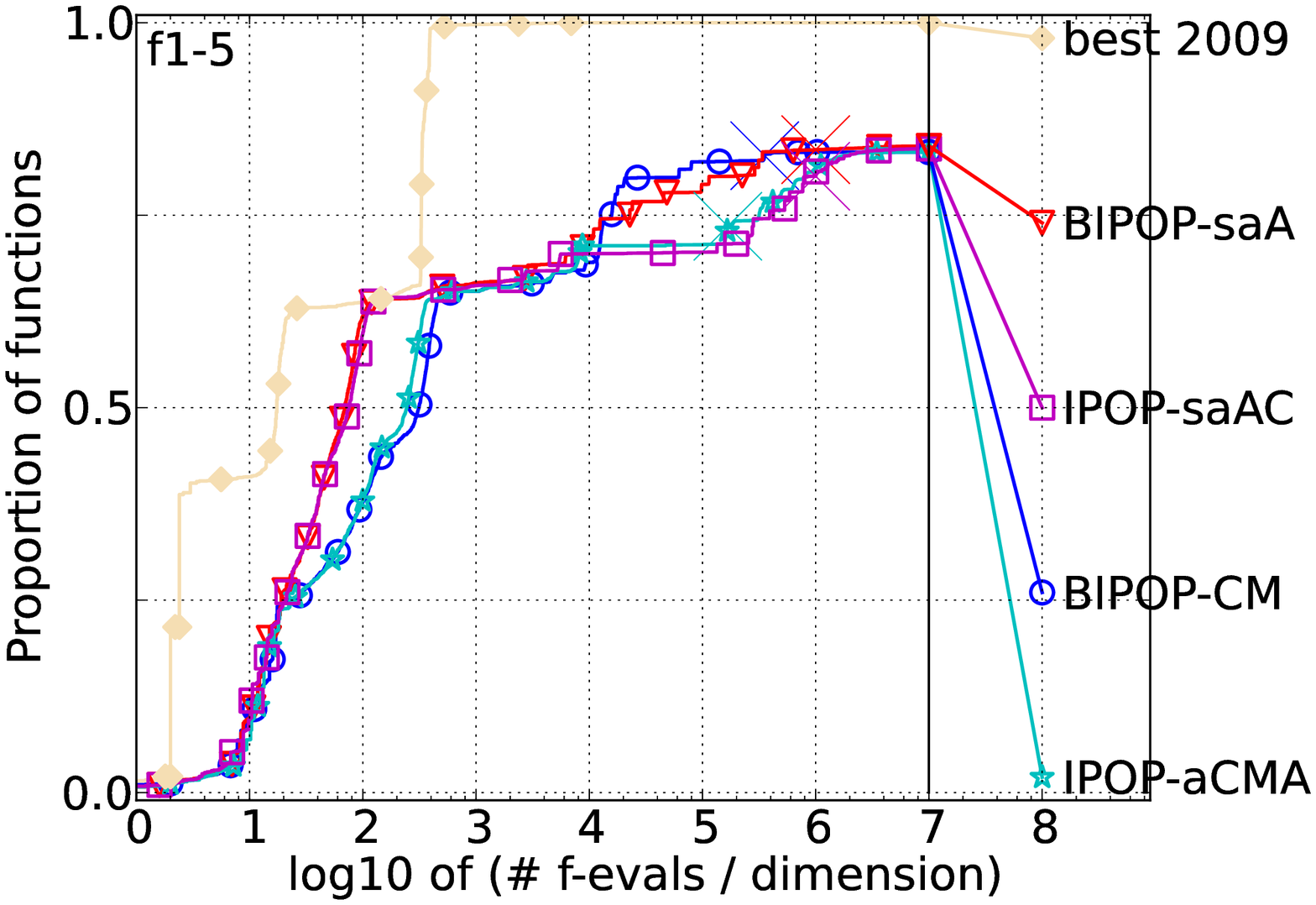}%
  \raisebox{.037\textwidth}{\parbox[b][.3\textwidth]{.0868\textwidth}{\begin{scriptsize}
    \perfprofsidepanel 
  \end{scriptsize}}}
 &
  \input{\bbobdatapath pprldmany_05D_lcond}%
  \includegraphics[width=0.4135\textwidth,trim=0mm 0mm 34mm 10mm, clip]{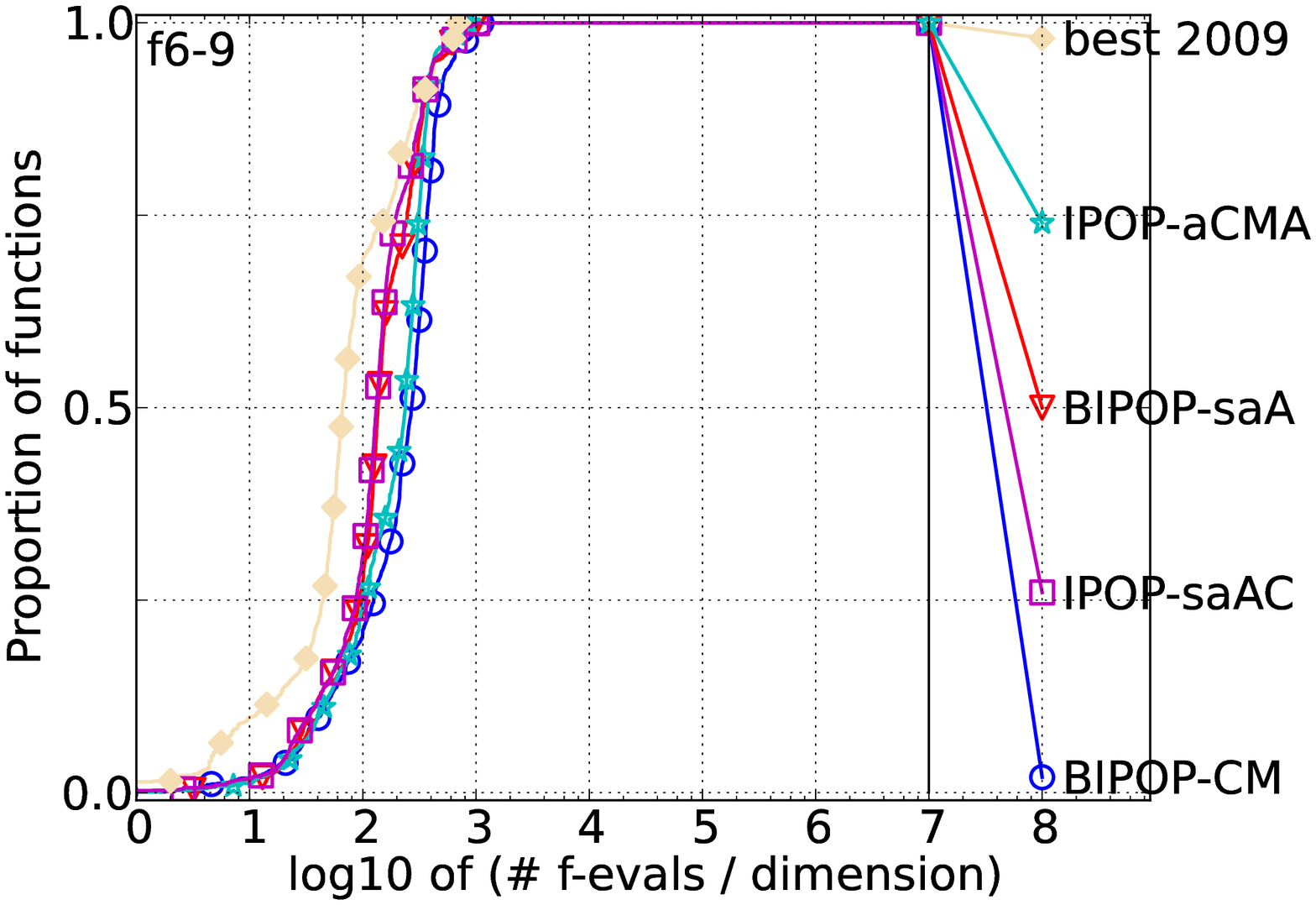}%
  \raisebox{.037\textwidth}{\parbox[b][.3\textwidth]{.0868\textwidth}{\begin{scriptsize}
    \perfprofsidepanel 
  \end{scriptsize}}}
 \\ 
ill-conditioned fcts & multi-modal fcts \\
  \input{\bbobdatapath pprldmany_05D_hcond}%
  \includegraphics[width=0.4135\textwidth,trim=0mm 0mm 34mm 10mm, clip]{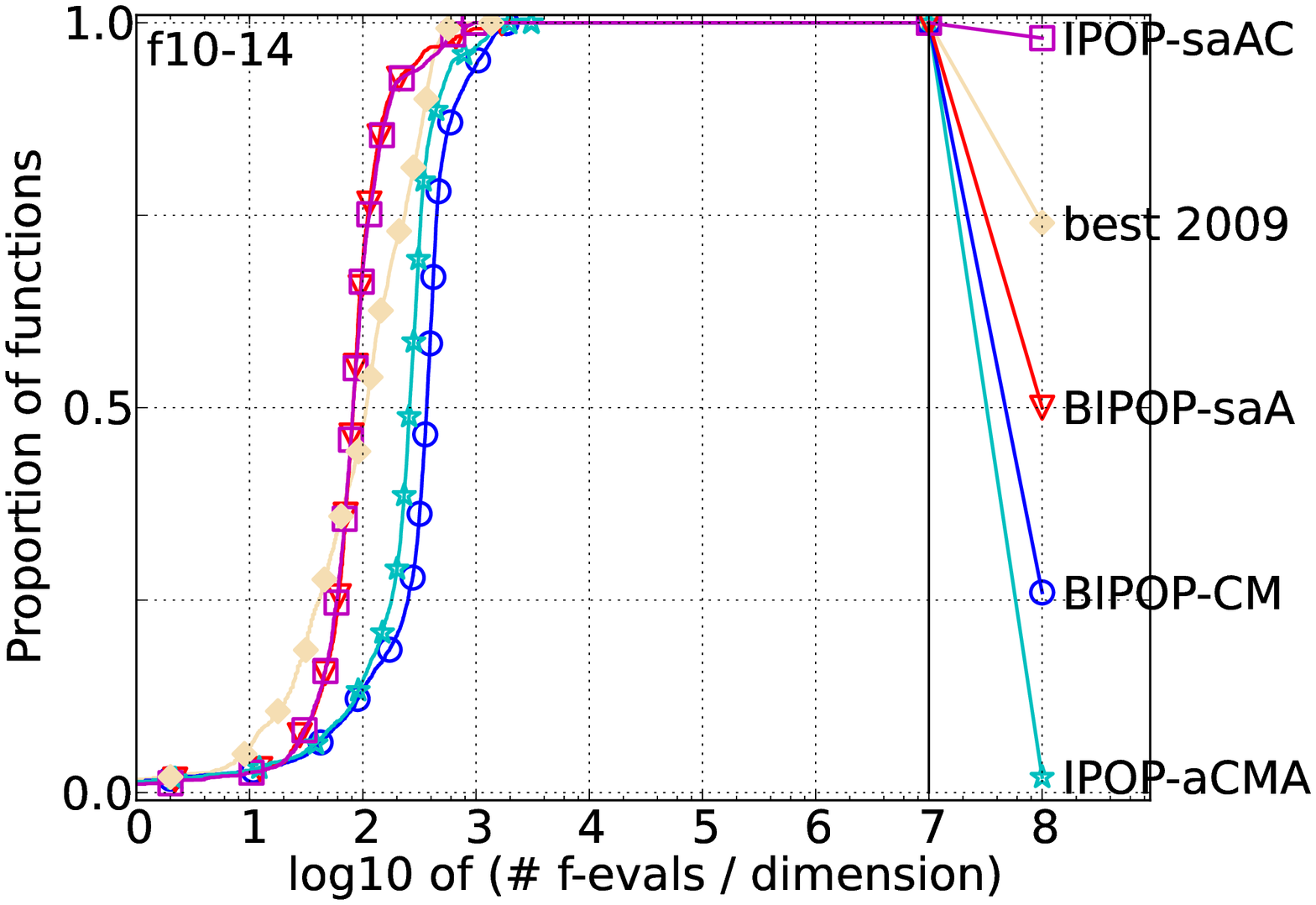}%
  \raisebox{.037\textwidth}{\parbox[b][.3\textwidth]{.0868\textwidth}{\begin{scriptsize}
    \perfprofsidepanel 
  \end{scriptsize}}}
 &
  \input{\bbobdatapath pprldmany_05D_multi}%
  \includegraphics[width=0.4135\textwidth,trim=0mm 0mm 34mm 10mm, clip]{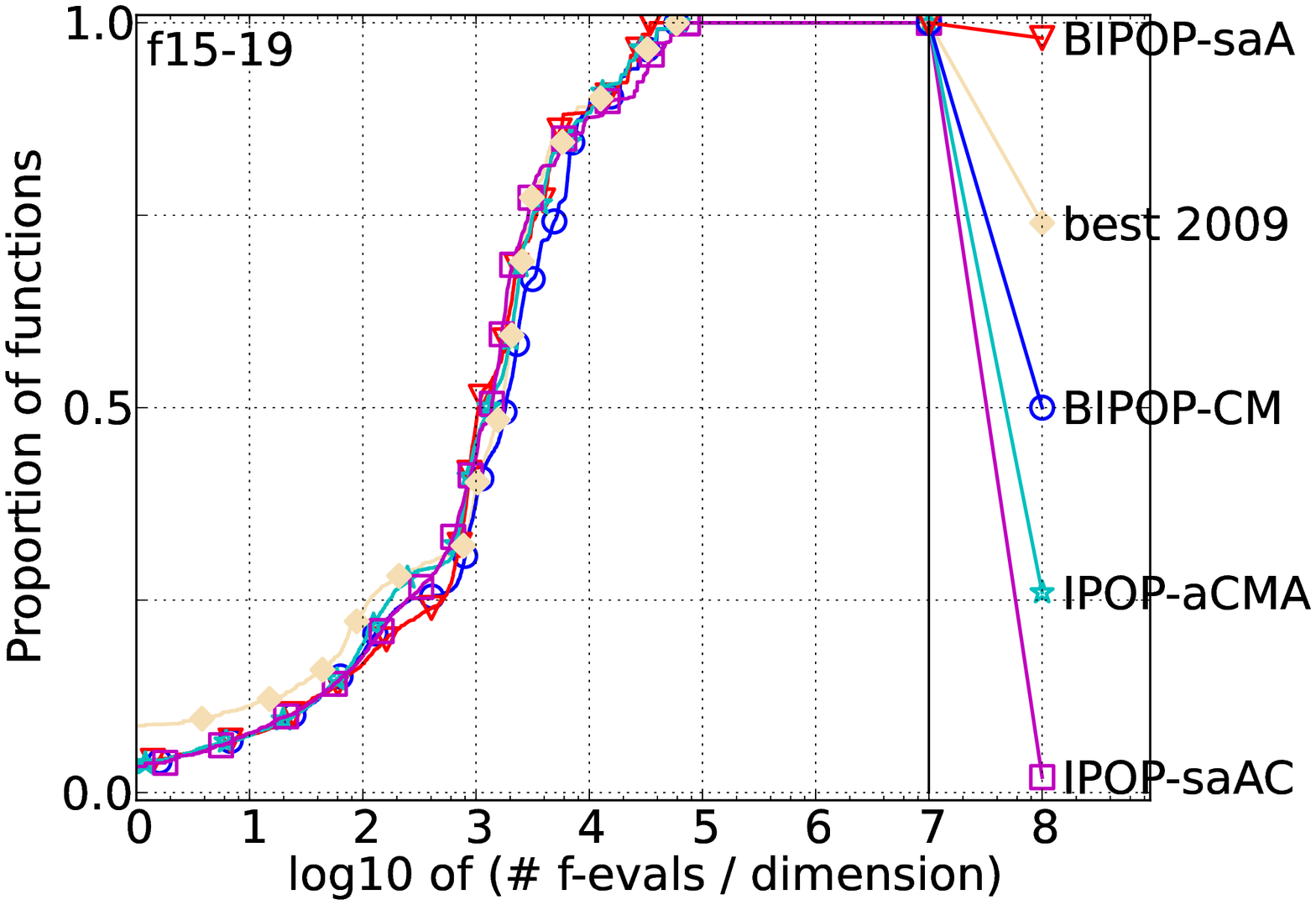}%
  \raisebox{.037\textwidth}{\parbox[b][.3\textwidth]{.0868\textwidth}{\begin{scriptsize}
    \perfprofsidepanel 
  \end{scriptsize}}}
 \\ 
 weakly structured multi-modal fcts & all functions\\
  \input{\bbobdatapath pprldmany_05D_mult2}%
  \includegraphics[width=0.4135\textwidth,trim=0mm 0mm 34mm 10mm, clip]{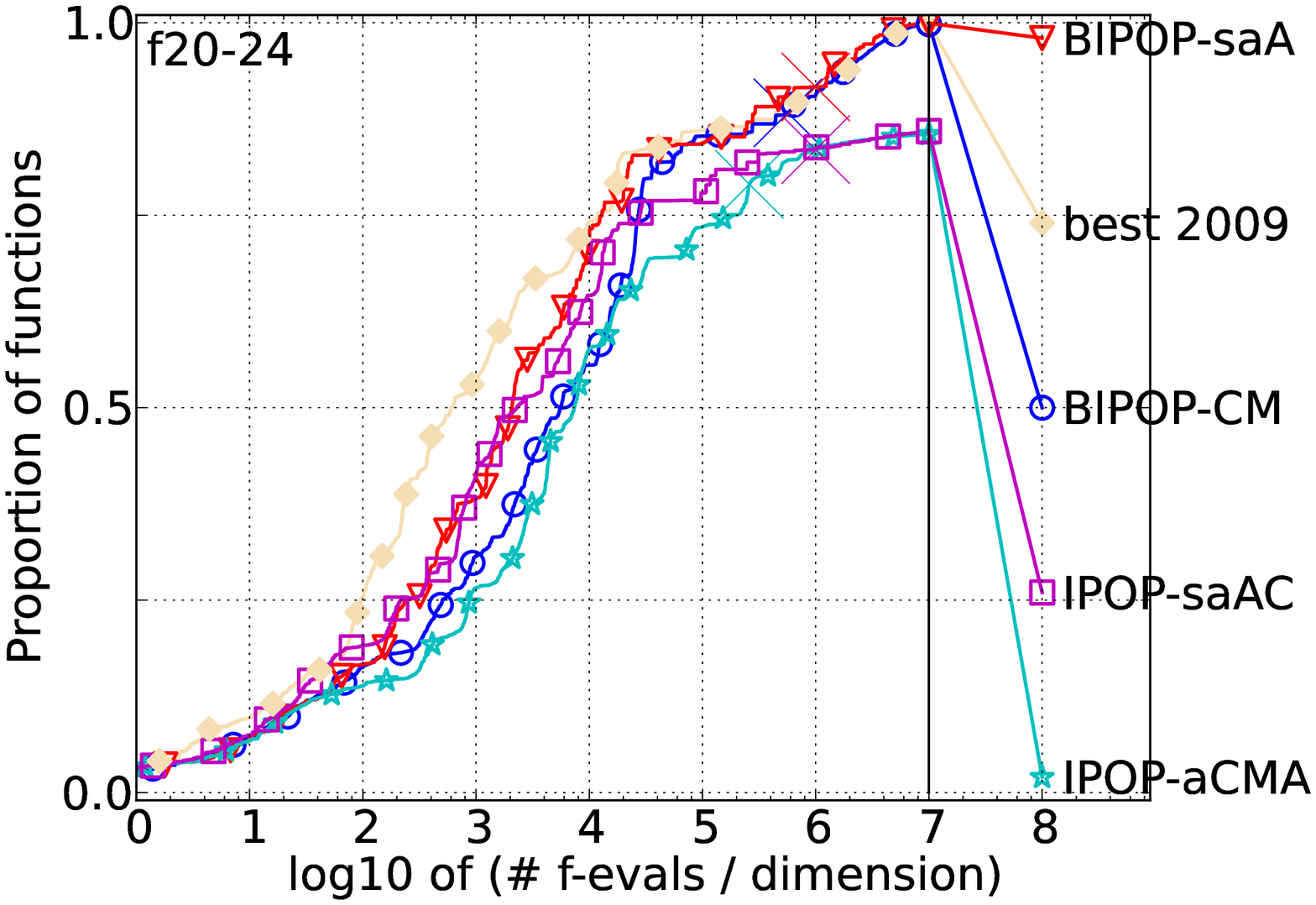}%
  \raisebox{.037\textwidth}{\parbox[b][.3\textwidth]{.0868\textwidth}{\begin{scriptsize}
    \perfprofsidepanel 
  \end{scriptsize}}}
 & 
  \input{\bbobdatapath pprldmany_05D_noiselessall}%
  \includegraphics[width=0.4135\textwidth,trim=0mm 0mm 34mm 10mm, clip]{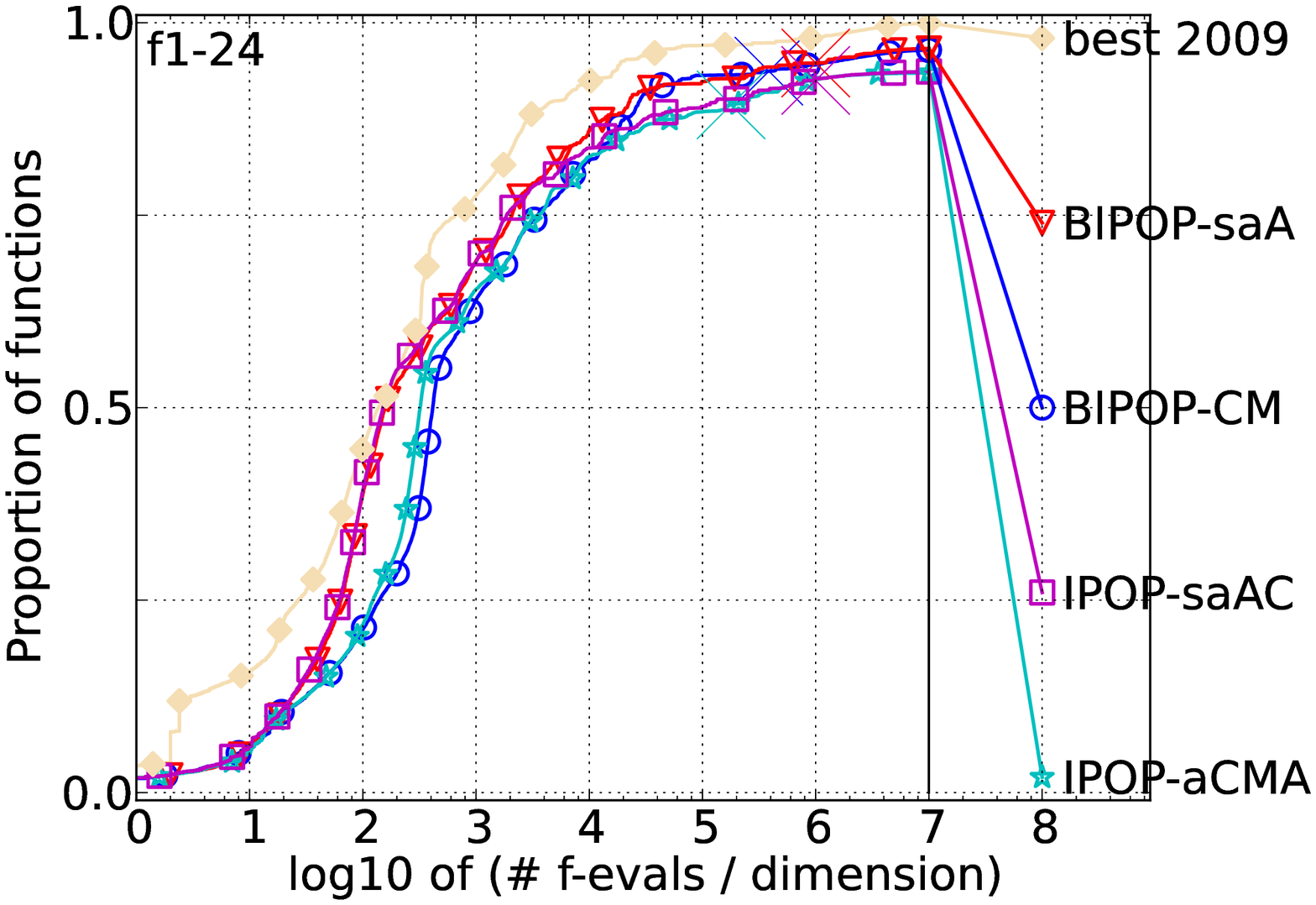}%
  \raisebox{.037\textwidth}{\parbox[b][.3\textwidth]{.0868\textwidth}{\begin{scriptsize}
    \perfprofsidepanel 
  \end{scriptsize}}}
 
 \end{tabular}
\caption{
\label{fig:ECDFs05D}
Bootstrapped empirical cumulative distribution of 
the number of objective function evaluations
divided by dimension (FEvals/D) for 50 targets in
$10^{[-8..2]}$ for all functions and subgroups in 5-D. The ``best 2009'' line
corresponds to the best \ERT\ observed during BBOB 2009 for each single target. 
}
\end{figure*}

\begin{figure*}
 \begin{tabular}{@{}c@{}c@{}}
 separable fcts & moderate fcts \\
  \input{\bbobdatapath pprldmany_20D_separ}%
  \includegraphics[width=0.4135\textwidth,trim=0mm 0mm 34mm 10mm, clip]{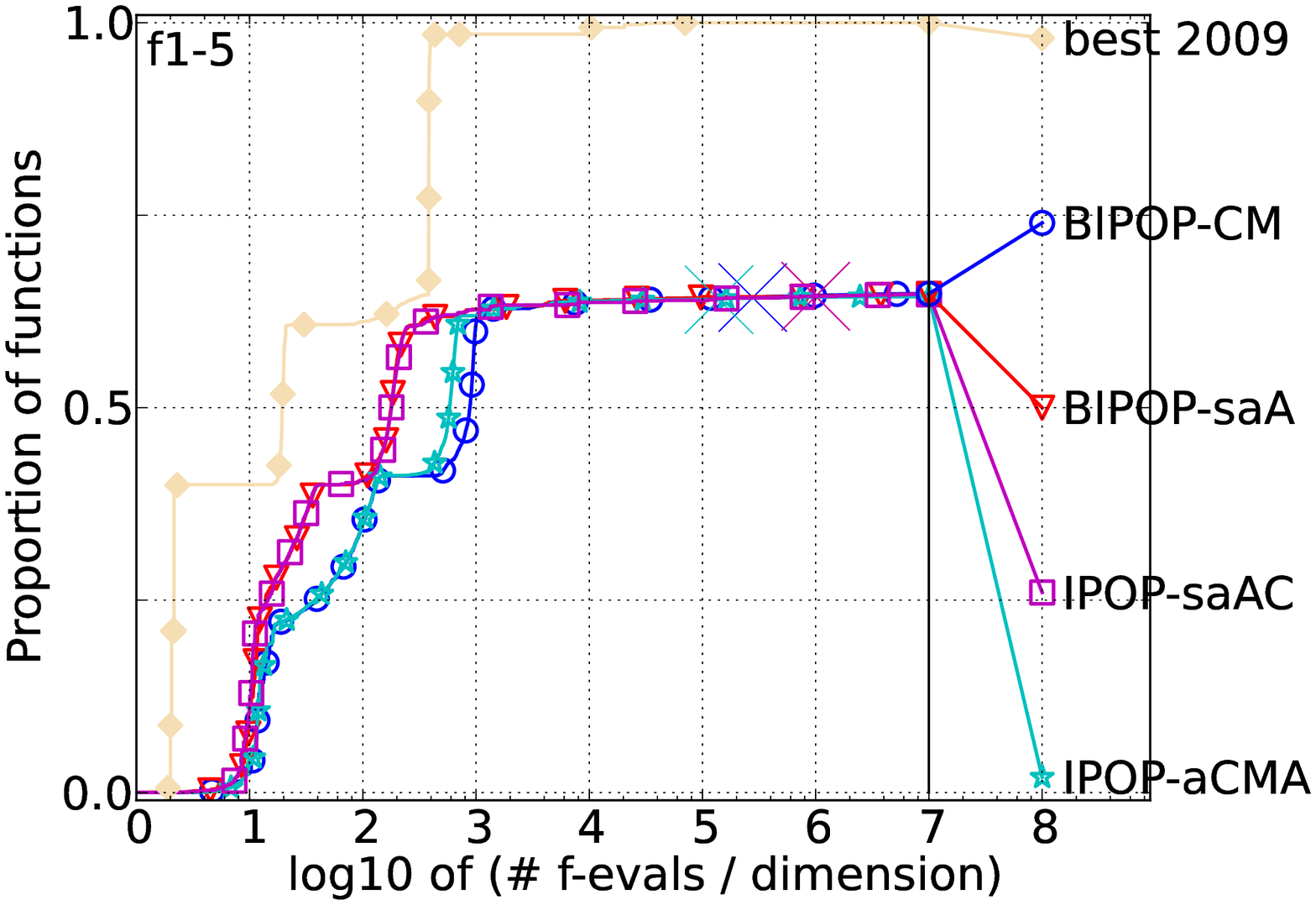}%
  \raisebox{.037\textwidth}{\parbox[b][.3\textwidth]{.0868\textwidth}{\begin{scriptsize}
    \perfprofsidepanel 
  \end{scriptsize}}}
 &
  \input{\bbobdatapath pprldmany_20D_lcond}%
  \includegraphics[width=0.4135\textwidth,trim=0mm 0mm 34mm 10mm, clip]{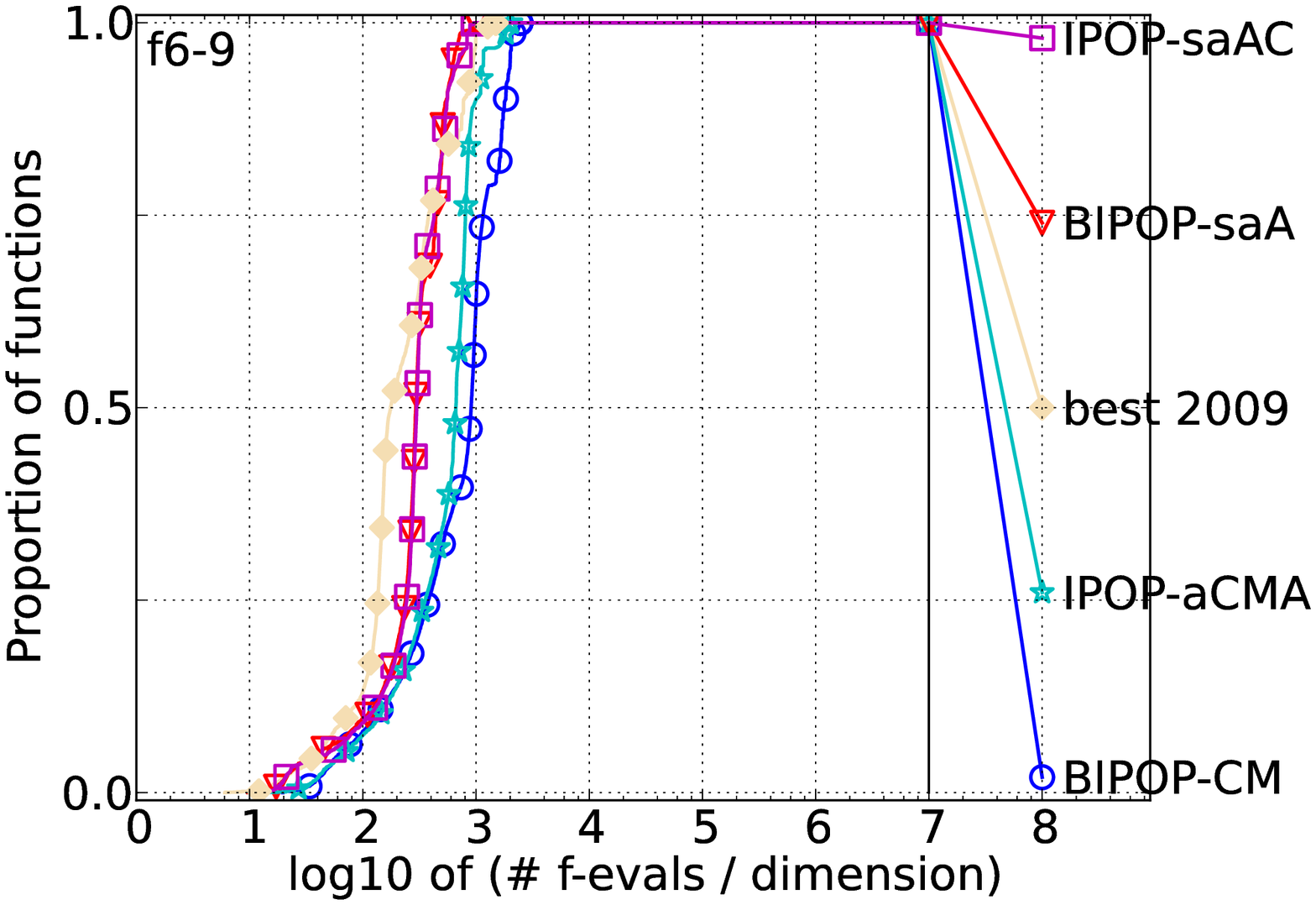}%
  \raisebox{.037\textwidth}{\parbox[b][.3\textwidth]{.0868\textwidth}{\begin{scriptsize}
    \perfprofsidepanel 
  \end{scriptsize}}}
 \\ 
ill-conditioned fcts & multi-modal fcts \\
  \input{\bbobdatapath pprldmany_20D_hcond}%
  \includegraphics[width=0.4135\textwidth,trim=0mm 0mm 34mm 10mm, clip]{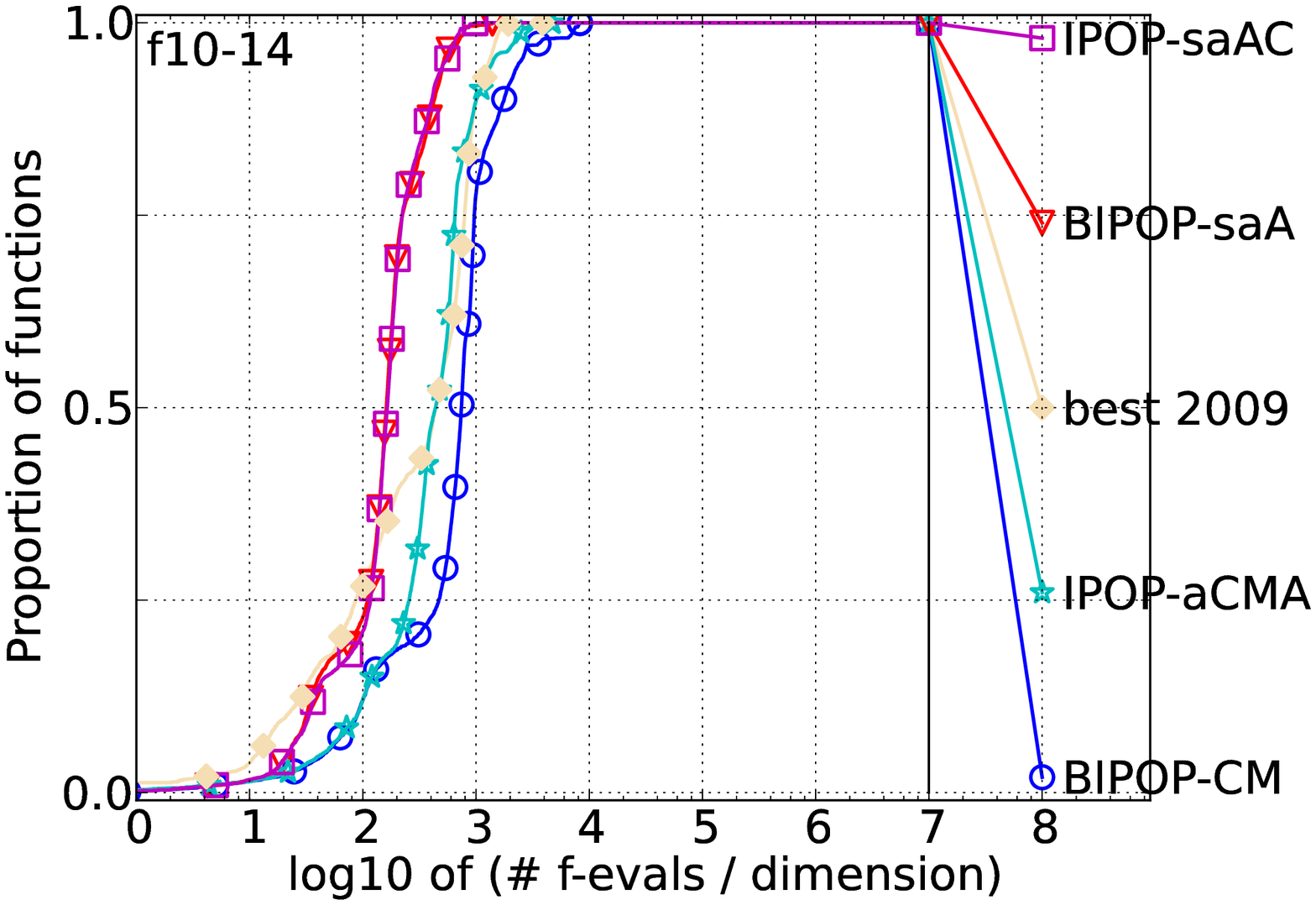}%
  \raisebox{.037\textwidth}{\parbox[b][.3\textwidth]{.0868\textwidth}{\begin{scriptsize}
    \perfprofsidepanel 
  \end{scriptsize}}}
 &
  \input{\bbobdatapath pprldmany_20D_multi}%
  \includegraphics[width=0.4135\textwidth,trim=0mm 0mm 34mm 10mm, clip]{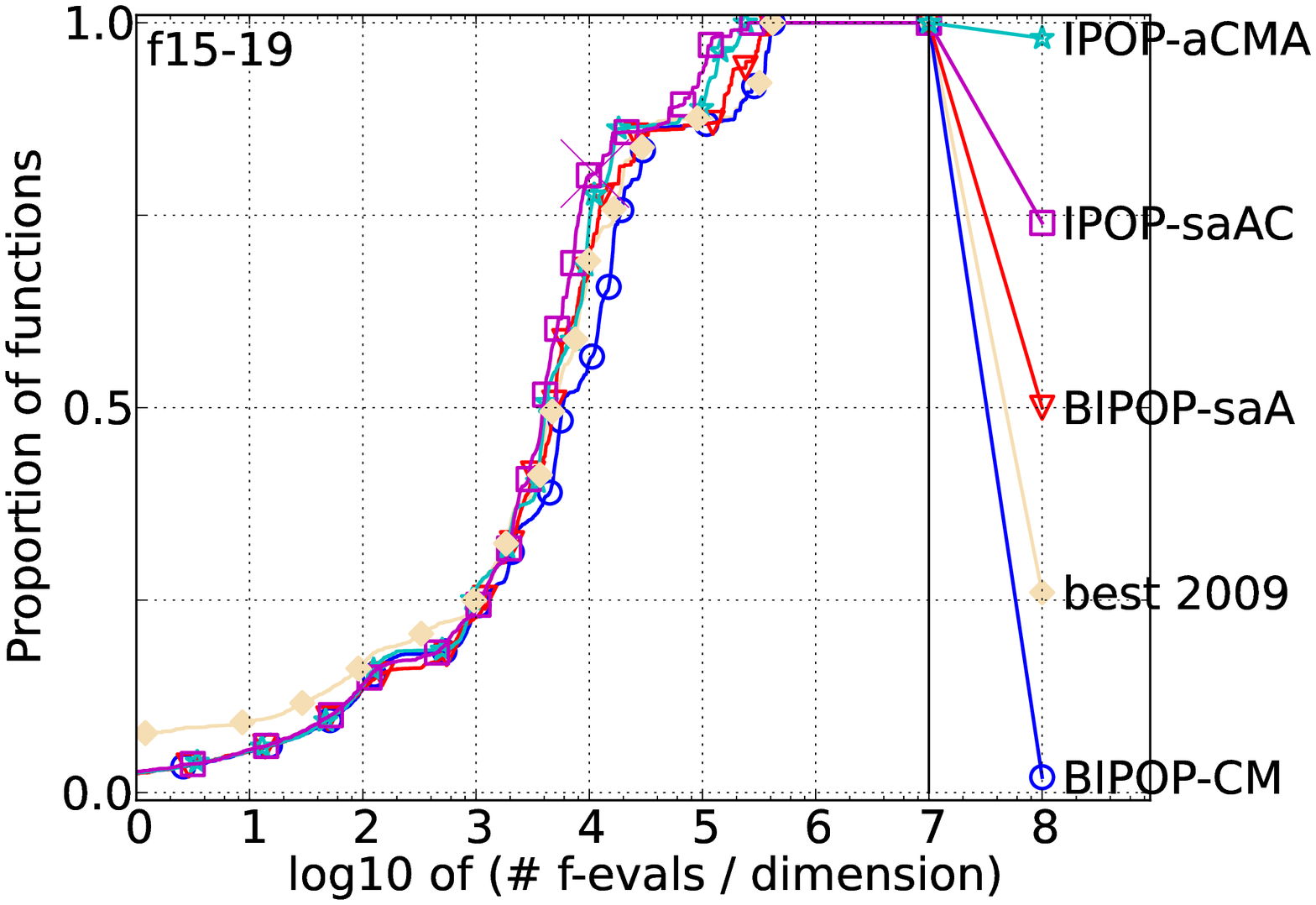}%
  \raisebox{.037\textwidth}{\parbox[b][.3\textwidth]{.0868\textwidth}{\begin{scriptsize}
    \perfprofsidepanel 
  \end{scriptsize}}}
 \\ 
 weakly structured multi-modal fcts & all functions\\
  \input{\bbobdatapath pprldmany_20D_mult2}%
  \includegraphics[width=0.4135\textwidth,trim=0mm 0mm 34mm 10mm, clip]{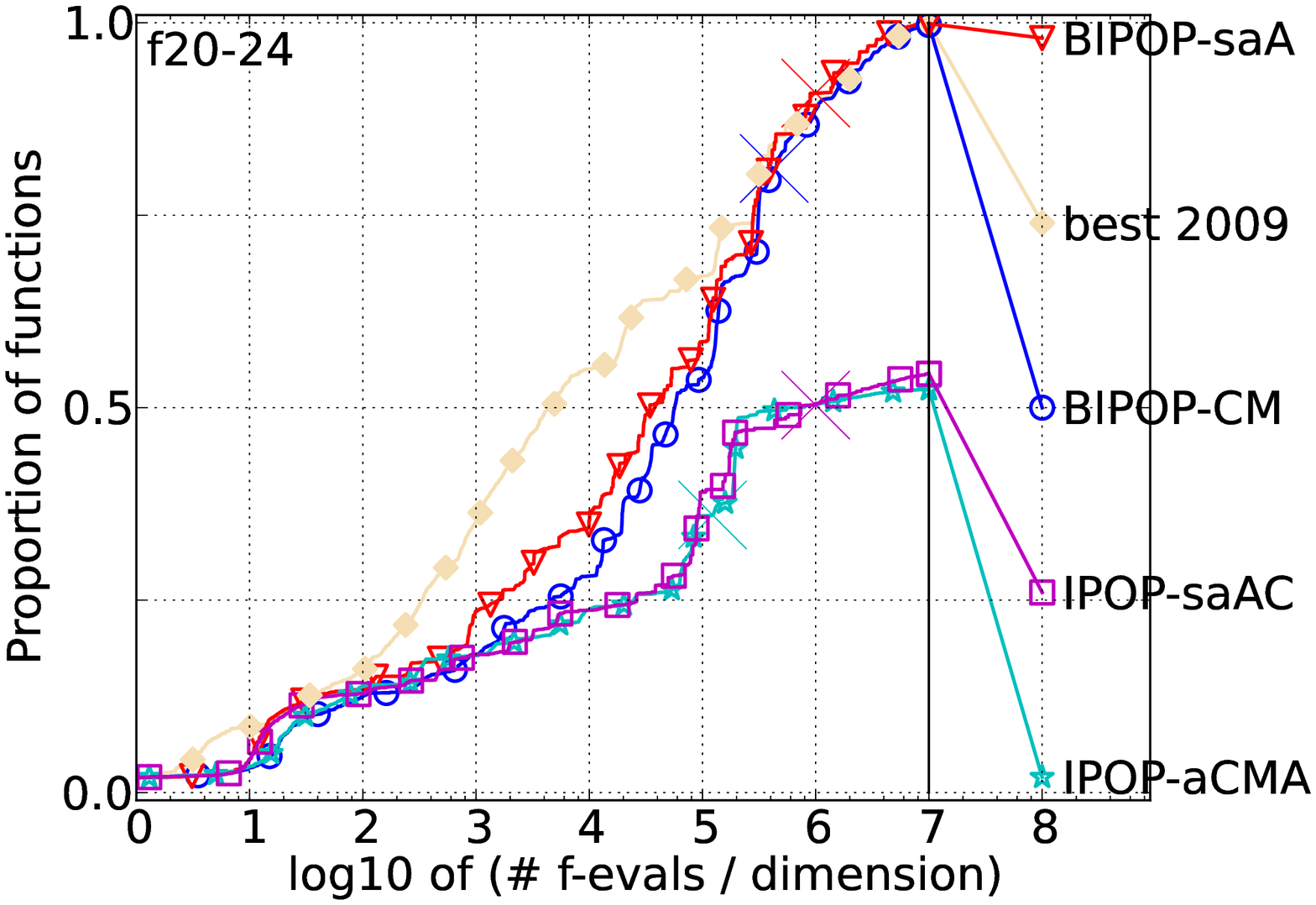}%
  \raisebox{.037\textwidth}{\parbox[b][.3\textwidth]{.0868\textwidth}{\begin{scriptsize}
    \perfprofsidepanel 
  \end{scriptsize}}}
 & 
  \input{\bbobdatapath pprldmany_20D_noiselessall}%
  \includegraphics[width=0.4135\textwidth,trim=0mm 0mm 34mm 10mm, clip]{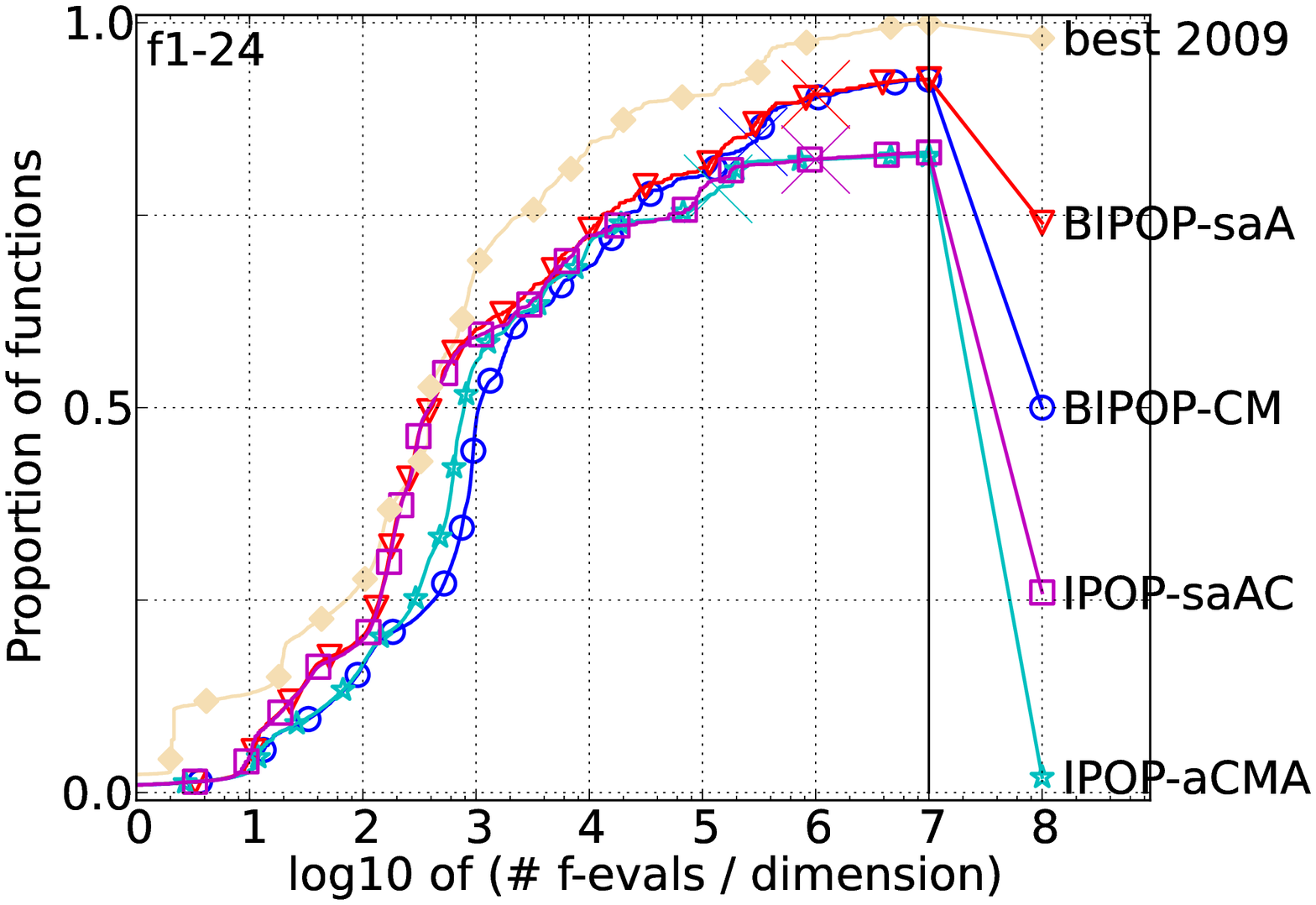}%
  \raisebox{.037\textwidth}{\parbox[b][.3\textwidth]{.0868\textwidth}{\begin{scriptsize}
    \perfprofsidepanel 
  \end{scriptsize}}}
 
 \end{tabular}
\caption{
\label{fig:ECDFs20D}
Bootstrapped empirical cumulative distribution of 
the number of objective function evaluations
divided by dimension (FEvals/D) for 50 targets in
$10^{[-8..2]}$ for all functions and subgroups in 20-D. The ``best 2009'' line
corresponds to the best \ERT\ observed during BBOB 2009 for each single target. 
}
\end{figure*}

Results from experiments according to \cite{hansen2012exp} on the
benchmark functions given in \cite{wp200901_2010,hansen2012fun} are
presented in Figures~\ref{fig:scaling}, \ref{fig:ECDFs05D} and
\ref{fig:ECDFs20D} and in Tables~\ref{tab:ERTs5} and~\ref{tab:ERTs20}. The
\textbf{expected running time (ERT)}, used in the figures and table,
depends on a given target function value, $\ftarget=\fopt+\Df$, and is
computed over all relevant trials (on the first 15 instances) as the number of function
evaluations executed during each trial while the best function value
did not reach \ftarget, summed over all trials and divided by the
number of trials that actually reached \ftarget\
\cite{hansen2012exp,price1997dev}.  \textbf{Statistical significance}
is tested with the rank-sum test for a given target $\Delta\ftarget$
($10^{-8}$ as in Figure~\ref{fig:scaling}) using, for each trial,
either the number of needed function evaluations to reach
$\Delta\ftarget$ (inverted and multiplied by $-1$), or, if the target
was not reached, the best $\Df$-value achieved, measured only up to
the smallest number of overall function evaluations for any
unsuccessful trial under consideration.

The \IPOPsaACM\ and \BIPOPsaACM\ represent the same algorithm (\saACM) before the first restart occurs, therefore, the results are very similar for the uni-modal functions, where the optimum usually can be found without restarts. 
The \saACM\ outperforms aCMA-ES usually by a factor from 2 to 4 on $f_1$,$f_2$,$f_8$,$f_9$$f_{10}$,$f_{11}$,$f_{12}$,$f_{13}$ and $f_{14}$ for dimensions between 5 and 20.
The speedup in dimension 2 is less pronounced for problems, where the running time is too short to improve the search. 
This is the case for $f_5$ Linear Slope function, where the speedup can be observed only for dimension 20, because the optimum can be found after about 200 function evaluations.
To improve the search on functions with small budgets it would make sense to use the surrogate model right after the first ($g_{start}=1$) generation of the CMA-ES, while in this study this parameter $g_{start}$ was set to 10 generations.

The good results on uni-modal functions can be explained by the fact, that while using the same amount of information (all previously evaluated points), 
\saACM\ processes this information in a more efficient way by constructing the approximation model of the function. Similar effect of more efficient exploitation of the available information can be observed for aCMA-ES in comparison to CMA-ES.

The speedup on multi-modal functions is less pronounced, because they are more difficult to approximate and the final surrogate model often has a bad precision.
In this case the adaptation of the number of generations leads to an oscillation of $\hat{n}$ close to 0, such that the surrogate model is not used for optimization or used for small number of generations. 

The BIPOP versions of CMA-ES usually perform better than IPOP on $f_{23}$ and $f_{24}$, where the optimum is more likely to be found if use small initial step-size. 
This leads to overall better performance of the BIPOP versions and \BIPOPsaACM\ in particular. The better performance of the latter in comparison with BIPOP-CMA-ES can be partially explained by the fact of using the active covariance matrix update. However, this is not the case for $f_{20}-f_{24}$ functions in 5-D and $f_{15-19}$ in 20-D (see Fig. \ref{fig:ECDFs05D} and Fig. \ref{fig:ECDFs20D}).

The \saACM\ algorithms improve the records in dimension 10 and 20 on $f_{7}$,$f_{10}$,$f_{11}$,$f_{12}$,$f_{13}$,$f_{14}$,$f_{15}$,$f_{16}$,$f_{20}$.

\section{CPU TIMING EXPERIMENT}

For the timing experiment the \IPOPsaACM\ was run on $f_1$, $f_8$, $f_{10}$ and $f_{15}$ without self-adaptation of surrogate model hyper-parameters. 
The crucial hyper-parameter for CPU time measurements, the number of training points was set $N_{training} =\left\lfloor 40+4D^{1.7}\right\rfloor$ as a function of dimension $D$.

These experiments have been conducted on a single core with 2.4 GHz under Windows XP using Matlab R2006a.

\begin{figure}[tb]
\begin{center}
  \includegraphics[scale=0.5]{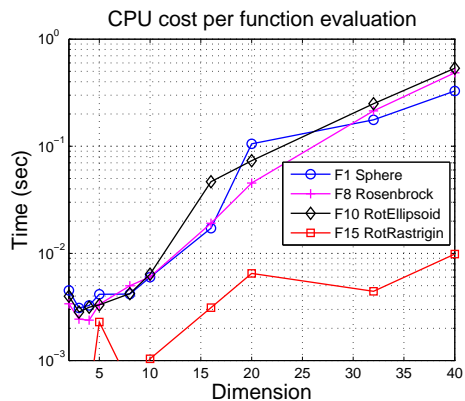}
\end{center}
\caption{\label{fig:time} CPU cost per function evaluation of IPOP-aACM-ES with fixed hyper-parameters.}
\end{figure}

On uni-modal functions the time complexity of surrogate model learning increases cubically in the search space dimension (see Fig. \ref{fig:time}) and quadratically in the number of training points. For small dimensions ($D<10$) the overall time complexity increases super-linearly in the dimension.
The time complexity per function evaluation depends on the population size, because one model is used to estimate the ranking of all points of the population. This leads to a smaller computational complexity on multi-modal functions, e.g. $f_{15}$ Rastrigin function, where the population becomes much larger after several restarts.

The results presented here does not take into account the model hyper-parameters optimization, where $\lambda_{hyp}$ surrogate models should be build at each iteration, which leads to an increase of CPU time per function evaluation by a factor of $\lambda_{hyp}$. For \BIPOPsaACM\ and \IPOPsaACM\ $\lambda_{hyp}$ was set to 20.

\section{Conclusion}

In this paper, we have compared the recently proposed self-adaptive surrogate-assisted \BIPOPsaACM\ and \IPOPsaACM\ with the BIPOP-CMA-ES and IPOP-aCMA-ES.
The surrogate-assisted \saACM\ algorithms outperform the original ones by a factor from 2 to 4 on uni-modal functions, and usually perform not worse on multi-modal functions. 
The \saACM\ algorithms improve the records on 8 out of 24 functions in dimension 10 and 20.

\section{ACKNOWLEDGMENTS}

The authors would like to acknowledge Anne Auger, Zyed Bouzarkouna, Nikolaus Hansen and Thomas P. Runarsson for their valuable discussions.
This work was partially funded by FUI of System@tic Paris-Region ICT cluster through contract DGT 117 407 {\em Complex Systems Design Lab} (CSDL).

\newcommand{\rot}[2][2.5]{
  \hspace*{-3.5\baselineskip}%
  \begin{rotate}{90}\hspace{#1em}#2
  \end{rotate}}
\newcommand{\includeperfprof}[1]{
  \input{\bbobdatapath #1}%
  \includegraphics[width=0.4135\textwidth,trim=0mm 0mm 34mm 10mm, clip]{#1}%
  \raisebox{.037\textwidth}{\parbox[b][.3\textwidth]{.0868\textwidth}{\begin{scriptsize}
    \perfprofsidepanel 
  \end{scriptsize}}}
}

\begin{table*}\tiny
\mbox{\begin{minipage}[t]{0.48\textwidth}\tiny
\centering
\input{\bbobdatapath pptables_f001_05D} 

\input{\bbobdatapath pptables_f002_05D}

\input{\bbobdatapath pptables_f003_05D}

\input{\bbobdatapath pptables_f004_05D}

\input{\bbobdatapath pptables_f005_05D}

\input{\bbobdatapath pptables_f006_05D}

\input{\bbobdatapath pptables_f007_05D}

\input{\bbobdatapath pptables_f008_05D}

\input{\bbobdatapath pptables_f009_05D}

\input{\bbobdatapath pptables_f010_05D}

\input{\bbobdatapath pptables_f011_05D}

\input{\bbobdatapath pptables_f012_05D}

\end{minipage}

\hspace{3mm}

\begin{minipage}[t]{0.48\textwidth}\tiny
\centering
\input{\bbobdatapath pptables_f013_05D}

\input{\bbobdatapath pptables_f014_05D}

\input{\bbobdatapath pptables_f015_05D}

\input{\bbobdatapath pptables_f016_05D}

\input{\bbobdatapath pptables_f017_05D}

\input{\bbobdatapath pptables_f018_05D}

\input{\bbobdatapath pptables_f019_05D}

\input{\bbobdatapath pptables_f020_05D}

\input{\bbobdatapath pptables_f021_05D}

\input{\bbobdatapath pptables_f022_05D}

\input{\bbobdatapath pptables_f023_05D}

\input{\bbobdatapath pptables_f024_05D}
\end{minipage}}

 \caption{\label{tab:ERTs5}
 \bbobpptablesmanylegend{dimension $5$}
 }
\end{table*}

\begin{table*}\tiny
\mbox{\begin{minipage}[t]{0.48\textwidth}\tiny
\centering
\input{\bbobdatapath pptables_f001_20D} 

\input{\bbobdatapath pptables_f002_20D}

\input{\bbobdatapath pptables_f003_20D}

\input{\bbobdatapath pptables_f004_20D}

\input{\bbobdatapath pptables_f005_20D}

\input{\bbobdatapath pptables_f006_20D}

\input{\bbobdatapath pptables_f007_20D}

\input{\bbobdatapath pptables_f008_20D}

\input{\bbobdatapath pptables_f009_20D}

\input{\bbobdatapath pptables_f010_20D}

\input{\bbobdatapath pptables_f011_20D}

\input{\bbobdatapath pptables_f012_20D}
\end{minipage}

\hspace{3mm}

\begin{minipage}[t]{0.48\textwidth}\tiny
\centering
\input{\bbobdatapath pptables_f013_20D}

\input{\bbobdatapath pptables_f014_20D}

\input{\bbobdatapath pptables_f015_20D}

\input{\bbobdatapath pptables_f016_20D}

\input{\bbobdatapath pptables_f017_20D}

\input{\bbobdatapath pptables_f018_20D}

\input{\bbobdatapath pptables_f019_20D}

\input{\bbobdatapath pptables_f020_20D}

\input{\bbobdatapath pptables_f021_20D}

\input{\bbobdatapath pptables_f022_20D}

\input{\bbobdatapath pptables_f023_20D}

\input{\bbobdatapath pptables_f024_20D}
\end{minipage}}
 \caption{\label{tab:ERTs20}
  \bbobpptablesmanylegend{dimension $20$}
}
\end{table*}

%
\bibliographystyle{abbrv}


%
%

\end{document}

%% file: ppdata/bbob_pproc_commands.tex
\providecommand{\bbobppfigsftarget}{\ensuremath{10^{-8}}}
\providecommand{\bbobppfigslegend}[1]{
Expected running time (\ERT\ in number of $f$-evaluations) divided by dimension for target function value \bbobppfigsftarget\ as $\log_{10}$ values versus dimension. Different symbols correspond to different algorithms given in the legend of #1. Light symbols give the maximum number of function evaluations from the longest trial divided by dimension. Horizontal lines give linear scaling, slanted dotted lines give quadratic scaling. Black stars indicate statistically better result compared to all other algorithms with $p<0.01$ and Bonferroni correction number of dimensions (six).  
Legend: 
{\color{blue}$\circ$}: \algorithmA
, {\color{red}$\triangledown$}: \algorithmB
, {\color{cyan}$\star$}: \algorithmC
, {\color{magenta}$\Box$}: \algorithmD
}
\providecommand{\bbobpptablesmanylegend}[1]{Expected running time (ERT in number of function evaluations)
                     divided by the respective best ERT measured during BBOB-2009 (given in 
                     the respective first row) for different $\Df$ values in #1. 
                     The central 80\% range divided by two is given in braces. 
                     The median number of conducted function evaluations is additionally given in 
                     \textit{italics}, if $\ERT(10^{-7}) = \infty$.
                     \#succ is the number of trials that reached the final target $\fopt + 10^{-8}$.
                     Best results are printed in bold. }
\providecommand{\algorithmA}{BIPOP-CMA}
\providecommand{\algorithmB}{BIPOP-saACM}
\providecommand{\algorithmC}{IPOP-aCMA}
\providecommand{\algorithmD}{IPOP-saACM}
\providecommand{\bbobppfigsftarget}{\ensuremath{10^{-8}}}
\providecommand{\bbobppfigslegend}[1]{
Expected running time (\ERT\ in number of $f$-evaluations) divided by dimension for target function value \bbobppfigsftarget\ as $\log_{10}$ values versus dimension. Different symbols correspond to different algorithms given in the legend of #1. Light symbols give the maximum number of function evaluations from the longest trial divided by dimension. Horizontal lines give linear scaling, slanted dotted lines give quadratic scaling. Black stars indicate statistically better result compared to all other algorithms with $p<0.01$ and Bonferroni correction number of dimensions (six).  
Legend: 
{\color{blue}$\circ$}: \algorithmA
, {\color{red}$\triangledown$}: \algorithmB
, {\color{cyan}$\star$}: \algorithmC
, {\color{magenta}$\Box$}: \algorithmD
}
\providecommand{\bbobpptablesmanylegend}[1]{Expected running time (ERT in number of function evaluations)
                     divided by the respective best ERT measured during BBOB-2009 (given in 
                     the respective first row) for different $\Df$ values in #1. 
                     The central 80\% range divided by two is given in braces. 
                     The median number of conducted function evaluations is additionally given in 
                     \textit{italics}, if $\ERT(10^{-7}) = \infty$.
                     \#succ is the number of trials that reached the final target $\fopt + 10^{-8}$.
                     Best results are printed in bold. }
\providecommand{\algorithmA}{BIPOP-CMA}
\providecommand{\algorithmB}{BIPOP-saACM}
\providecommand{\algorithmC}{IPOP-aCMA}
\providecommand{\algorithmD}{IPOP-saACM}
\providecommand{\algorithmA}{BIPOP-CMA}
\providecommand{\algorithmB}{BIPOP-aCMA}
\providecommand{\algorithmC}{BIPOP-saACM}
\providecommand{\algorithmD}{IPOP-aCMA}

\providecommand{\bbobppfigsftarget}{\ensuremath{10^{-8}}}
\providecommand{\bbobppfigslegend}[1]{
Expected running time (\ERT\ in number of $f$-evaluations) divided by dimension for target function value \bbobppfigsftarget\ as $\log_{10}$ values versus dimension. Different symbols correspond to different algorithms given in the legend of #1. Light symbols give the maximum number of function evaluations from the longest trial divided by dimension. Horizontal lines give linear scaling, slanted dotted lines give quadratic scaling. Black stars indicate statistically better result compared to all other algorithms with $p<0.01$ and Bonferroni correction number of dimensions (six).  
Legend: 
{\color{blue}$\circ$}: \algorithmA
, {\color{red}$\triangledown$}: \algorithmB
, {\color{cyan}$\star$}: \algorithmC
, {\color{magenta}$\Box$}: \algorithmD
}
\providecommand{\bbobpptablesmanylegend}[1]{Expected running time (ERT in number of function evaluations)
                     divided by the respective best ERT measured during BBOB-2009 (given in 
                     the respective first row) for different $\Df$ values in #1. 
                     The central 80\% range divided by two is given in braces. 
                     The median number of conducted function evaluations is additionally given in 
                     \textit{italics}, if $\ERT(10^{-7}) = \infty$.
                     \#succ is the number of trials that reached the final target $\fopt + 10^{-8}$.
                     Best results are printed in bold. }
\providecommand{\algorithmA}{BIPOP-CMA}
\providecommand{\algorithmB}{BIPOP-saACM}
\providecommand{\algorithmC}{IPOP-aCMA}
\providecommand{\algorithmD}{IPOP-saACM}